\documentclass[sigconf,nonacm]{acmart}

\AtBeginDocument{%
  \providecommand\BibTeX{{%
    \normalfont B\kern-0.5em{\scshape i\kern-0.25em b}\kern-0.8em\TeX}}}

\usepackage{dsfont}
\usepackage{amsmath}
\usepackage{color}
\usepackage[normalem]{ulem}
\usepackage{tabularray}
\usepackage{multirow}
\usepackage{subcaption}
\usepackage{algorithm}  
\usepackage[noend]{algpseudocode}
\usepackage[english]{babel}
\newtheorem{theorem}{Theorem}
\usepackage[normalem]{ulem}
\useunder{\uline}{\ul}{}
\usepackage{pifont}%
\usepackage{tabularray}
\setcopyright{acmcopyright}
\copyrightyear{2023}
\acmYear{2023}
\acmDOI{XXXXXXX.XXXXXXX}

\acmConference[WebConf '24]{The Web Conference}{May 13 - 17, 2024}{Singapore}
\acmPrice{15.00}
\acmISBN{978-1-4503-XXXX-X/05/2024}

\begin{document}

\title{Are We Wasting Time? A Fast, Accurate Performance Evaluation Framework for Knowledge Graph Link Predictors}
\newif\ifanonymize
\anonymizefalse  %

\ifanonymize
  \author{Anonymous Author(s)}
\else
\author{Filip Cornell}
\authornote{Corresponding author.}
\email{fcornell@kth.se}
\affiliation{%
  \institution{KTH Royal Institute of Technology \& Gavagai}
  \city{Stockholm}
  \country{Sweden}
  \postcode{11224}
}

\author{Yifei Jin}
\email{yifeij@kth.se}
\affiliation{%
  \institution{KTH Royal Institute of Technology \& Ericsson}
  \city{Stockholm}
  \country{Sweden}}

\author{Jussi Karlgren}
\email{jussi@lingvi.st}
\affiliation{%
 \institution{Silo AI}
 \city{Helsinki}
 \country{Finland}}

\author{Sarunas Girdzijauskas}
\email{sarunasg@kth.se}
\affiliation{%
 \institution{KTH Royal Institute of Technology}
  \streetaddress{30 Shuangqing Rd}
  \city{Stockholm}
  \country{Sweden}}

\fi
\renewcommand{\shortauthors}{Cornell et al.}

\begin{abstract}
  The standard evaluation protocol for measuring the quality of Knowledge Graph Completion methods --- the task of inferring new links to be added to a graph --- typically involves a step which ranks every entity of a Knowledge Graph to assess their fit as a head or tail of a candidate link to be added. In Knowledge Graphs on a larger scale, this task rapidly becomes prohibitively heavy. Previous approaches mitigate this problem by using random sampling of entities to assess the quality of links predicted or suggested by a method. However, we show that this approach has serious limitations since the ranking metrics produced do not properly reflect true outcomes. In this paper, we present a thorough analysis of these effects along with the following findings. First, we empirically find and theoretically motivate why sampling uniformly at random vastly overestimates the ranking performance of a method. We show that this can be attributed to the effect of $\textit{easy}$ versus $\textit{hard}$ negative candidates. Second, we propose a framework that uses relational recommenders to guide the selection of candidates for evaluation. We provide both theoretical and empirical justification of our methodology, and find that simple and fast methods can work extremely well, and that they match advanced neural approaches. Even when a large portion of the true candidates for a property are missed, the estimation of the ranking metrics on a downstream model barely deteriorates. With our proposed framework, we can reduce the time and computation needed similar to random sampling strategies while vastly improving the estimation; on ogbl-wikikg2, we show that accurate estimations of the full, filtered ranking can be obtained in 20 seconds instead of 30 minutes. We conclude that considerable computational effort can be saved by effective preprocessing and sampling methods and still reliably predict performance accurately of the true performance for the entire ranking procedure.

\end{abstract}




\received{20 February 2007}
\received[revised]{12 March 2009}
\received[accepted]{5 June 2009}

\maketitle
\begin{figure}
  \includegraphics[width=\linewidth]{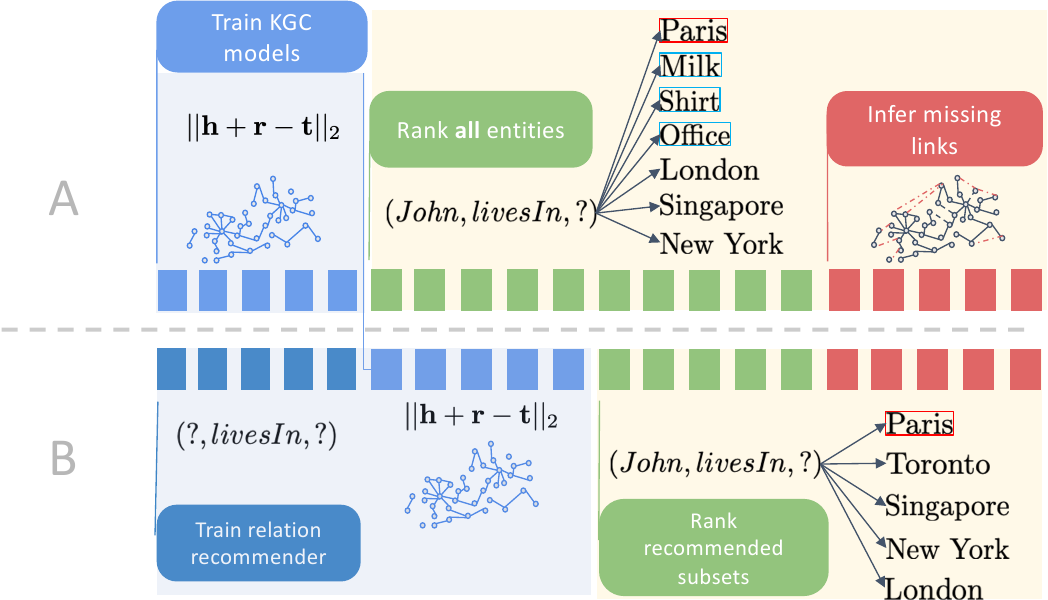}
  \caption{\textbf{A}: Normally, one ranks all, even highly irrelevant entities during evaluation. \textbf{B}: Instead, we propose using small, accurate subsets to alleviate the heavy burden of evaluation. }
  \label{fig:make-sarunas-happy-fig}
\end{figure}
\section{Introduction}

Knowledge Graphs (KGs) have proven a useful knowledge representation for a range of practical tasks across domains, such as Question Answering \cite{lukovnikov2017neural, bao2016constraint}, Retrieval-Augmented Generation \cite{yu2022retrieval} and other tasks. Knowledge Graphs have settled into their current form in the past decades but trace their roots to various explicit semantic representations such as semantic nets, conceptual models, and various instances of category theory. %

As Knowledge Graphs are notoriously incomplete \cite{chen2020knowledge} missing links need to be added and since Knowledge Graphs often are of considerable scale, \cite{vrandevcic2014wikidata, auer2007dbpedia, yago}, preserving the quality of a Knowledge Graph becomes a notable challenge for knowledge engineers and editors. Knowledge Grapch Completion (KGC) models can be developed to make the addition of missing links a semi- or fully automatic process. Typically, this is formulated and evaluated as a ranking problem \cite{bordes2013translating}: for each head-relation pair $(h,r, ?)$ and each relation-tail pair $(?, r, t)$, every entity in the graph is ranked as a candidate for the missing tail or head and then the ranking of the true entity, where one is known, can be used as a metric for how well a KGC method predicts missing items. This results in $|\mathcal{E}|$ operations for each ranking query where $\mathcal{E}$ is the set of entities. As the KG grows larger, this becomes a very expensive and impracticable operation; for a complete ranking evaluation, the complexity grows at the rate of $\mathcal{O}(|\mathcal{E}|^2)$ \cite{bastos2023can}. To handle this at scale, previous efforts have resorted to using approximations. Some use a uniform random sample set of counterexamples ("negative") entities and compute the ROC-AUC metric \cite{hajian2013receiver} instead of a full evaluation \cite{teru2020inductive}. This is highly effective in reducing the workload of the evaluation, as the reduction of time needed also decreases at the rate of $|\mathcal{E}|^2$. For example, on the ogbl-wikikg2 benchmark \cite{hu2021ogblsc}, a large-scale subset of Wikidata \cite{vrandevcic2014wikidata}, the ranking is done by ranking $1,000$ pre-defined candidate heads and tails ($500$ of each), and the results are reported as an average and standard deviation of the Mean Reciprocal Rank (MRR). While this indeed alleviates the computational effort, such a sampling approach fails to capture the genuine distribution of the entity data, potentially limiting the accuracy of insights into how well a Knowledge Graph Completion method under consideration performs. 

In this study, we find that these sampled estimations are overly optimistic for estimating ranking metrics, and full-ranking metric results are significantly, up to 40 percentage points, lower than reported. We demonstrate that this can be attributed to \textit{hard} and \textit{easy} negatives, as most sampled entities end up semantically irrelevant to the relation in the query. We propose a framework to solve this by utilizing Relation Recommenders (see Figure \ref{fig:make-sarunas-happy-fig}), a more general version of Property Recommender methods \cite{zangerle2016empirical}; i.e., methods defining scores for an entity being a head or a tail of a relation $r$. With these, we form \textit{small} head- and tail sets, known as the \textit{domains} \& \textit{ranges} in an ontology \cite{hogan2021knowledge} that are used to sample more difficult candidates, mitigating the over-estimation caused by a large number of easy false answers by skipping them and focusing and more difficult examples. To our knowledge, this is the first study that explores the impact of incorporating domain and range knowledge to streamline the running cost, evaluation time, and complexity of KGC model evaluation. Our framework is model-agnostic and can work with any KGC model, allowing for quicker development and iterations as well as accurate evaluation estimates. As this set of methods only considers the relation for a query and is agnostic to the entities, we limit the expensive process of sampling to the order of number of relationship types $|\mathcal{R}|$ for the evaluation instead of sampling for each query $(h, r, ?)$. To summarize, our main contributions are:

\begin{enumerate}
    \item We demonstrate that sampling candidates guided by relation recommender model scores drastically reduces the time needed to evaluate a KG Completion model on large-scale datasets and obtains accurate estimations of ranking metrics. On ogbl-wikikg2, a dataset with millions of entities, our framework allows us to use only $2 \%$ of all entities while accurately estimating the true ranking metrics, running in approximately $20$ seconds instead of $30$ minutes, exhibiting a $90$-fold improvement.
    \item We provide both theoretical and empirical justifications for our methods, and extensive experiments show that our method outperforms baselines both in terms of correlation and absolute estimation of the value allowing an accurate estimation of ranking metrics. 
    \item We compare relational recommenders based on a number of criteria; not only their performance, but also scalability, ease-of-use and generality, providing an overarching view on which workload different relational recommenders are appropriate.
\end{enumerate}

\section{Related work} \label{sec:relatedwork}

\paragraph{Evaluating Knowledge Graph Completion} Most KGC experiments evaluate using the same set-up, with Filtered Mean Reciprocal Rank (MRR) as well as Hits@X, with X usually being $1$, $3$, $5$ or $10$ \cite{bordes2013translating}. Others have considered measures such as Precision and Recall \cite{Speranskaya2020RankingVC}, and in recent inductive KGC experiments, the ROC-AUC based on sampling various times have been used \cite{teru2020inductive}. Rim et al. \cite{rim2021behavioral} evaluate a model's capability on its ability to model different relation properties such as symmetry, hierarchy, and other capabilities. Various other works \cite{mohamed2020popularity, berrendorf2020interpretable} have considered different aspects of the evaluation of KGC models; however, not a lot of work has been done on reducing the demands of the evaluation in KG Completion. The work closest to ours is the Knowledge Persistence ($\mathcal{KP}$) framework introduced by Bastos et al. \cite{bastos2023can}. They show that Persistent Homology can provide an evaluation metric that correlates with the ranking metric. By sampling two weighted, directed graphs, $\mathcal{KP}^+$ and $\mathcal{KP}^-$ of positive and negative triples and computing the Persistent Homology, $\mathcal{KP}$ extracts information about the geometry of the latent space. By taking the Sliced Wasserstein distance between the Persistence Diagrams of $\mathcal{KP}^+$ and $\mathcal{KP}^-$, they derive a metric as to how similar the two graphs are. Through their measure, they manage to derive a metric that they show correlates with the ranking metrics and is significantly faster as they remain on the computational complexity of $\mathcal{O}(|\mathcal{E}|)$ instead of $\mathcal{O}(|\mathcal{E}|^2)$.

Similar to other approximations such as random sampling \cite{hu2020open}, this method risks being biased towards easier examples. Moreover, they develop a new metric that highly correlates with the ranking metrics. This makes it harder to estimate and evaluate the performance of an individual query which limits opportunities for interpretability and analysis. We show below in Section \ref{sec:results} that the correlation of $\mathcal{KP}$ varies across datasets and models, with no guarantee for a strong correlation. 

In this presented approach we maintain the objective to estimate individual ranks and then aggregate those to exact metrics. These proposed approaches can be combined, since sampling negative triplet uniformly at random can be boosted by adding samples of harder candidates.

\paragraph{Candidate generation through constructing domains and ranges} More generally, candidate generation prior to ranking has been adopted in many recommender systems \cite{medvedev2019powered,gupta2021recpipe,kang2019candidate} but less so in KG completion. Several works \cite{safavi-koutra-2020-codex,wang-etal-2022-simkgc} have addressed this implicitly by discussing hard versus easy negatives; for example, Safavi et al. \cite{safavi-koutra-2020-codex} creating the CoDeX datasets show that triplet classification on random samples is an easy, nearly solved task; however, classifying annotated, more credible triples is significantly more difficult. 

Borrego et al. \cite{borrego2019generating} create CHAI, a method considering both the head and the relation to generate tail candidates for KGs with a small number of relations. Similarly, Christmann et al. \cite{christmann2022beyond} focus on reducing the search space specifically for specific queries over Knowledge Graphs through Named Entity Disambiguation, adopting the modality of text which is not necessarily available or of high quality in Knowledge Graphs. Dubey et al. \cite{dubey2018earl} reduce the search space for joint entity- and relation linking, and also take a textual approach. The above-mentioned methods focus however on candidate generation for queries in the format $(h, r, ?)$ and therefore consider which head is part of the query. We instead focus on leveraging methods agnostic to the entities involved, i.e., $(\cdot, r, ?)$.

The task of defining head and tail sets, i.e., domains and ranges has been addressed by previous research and has gone under various different names. One way of modelling this has been methods focusing on predicting properties of entities \cite{abedjan2014amending, chao2022pie, gassler2014guided}, although often focused purely on tail sets. The Wikidata Property Suggester (shortened WD by Zangerle et al. \cite{zangerle2016empirical}, an abbreviation we also adopt) \footnote{\url{https://gerrit.wikimedia.org/r/admin/repos/mediawiki\%2Fextensions\%2FPropertySuggester,general}}, based on the work created by Abedjan et al. \cite{abedjan2014amending} functions as a recommender system for WikiData creators to suggest potential properties of users. However, none of these focus on the \textit{inverse} relations. Zangerle et al. \cite{zangerle2016empirical} compare several different property recommenders such as Snoopy \cite{gassler2014guided} and WD, and found WD to work the best. 

One of the most popular methods for heuristically defining has been what is referred to by the \textit{Pykeen} \cite{ali2021pykeen} creators as the \textit{PseudoTyped} (\textbf{PT}) method, adopted in many previous works to mainly improve sampling in KGC \cite{balkir-etal-2019-using,balkir2018improving,krompass2015type,shi2018open}. However, this method comes with a major drawback; it cannot generate candidates previously not seen in a domain or range. This is detrimental for relations such as \textit{1-1}, \textit{1-M} (one-to-many) and \textit{M-1} (many-to-one) relations, where a relation only exists once per entity; for example \verb|isMarriedTo|. In these cases, a PT approach is unable to find the correct candidates and the method therefore poses severe limitations. 

In the large-scale (LSC) OGB competition \cite{hu2021ogblsc}, two teams performed candidate generation by defining domains \& ranges. Chen et al. \cite{chensolution} implemented a Degree-Based Heuristic (DBH). The method scores each entity's likelihood of being a head of a relation by the occurrence as the number of times it occurs as a head of the relation, and respectively for the tail. For example, if the entity \verb|France| has been seen in the training set as a tail for the \verb|countryOfOrigin| 1,000 times, its score will be 1,000. This method is upper-bounded in terms of Candidate Recall by \textbf{PT}, and therefore also suffering from the limitations that it cannot see unseen relational candidates For the remainder of this paper, we will therefore compare to PT rather than DBH, as it is upper-bounded in terms of recall by PT. As a complement to DBH, Chen et al. considered a rule-based approach for generating candidates. This was however not constructing domains \& ranges, but rather a form of KGC as the entity was needed and can be considered a KGC method, as it requires knowing both which head and its relation to score. 

Chao et al. \cite{chao2022pie} built a solution based on a previous model called PathCon \cite{wang2021relational}, but instead of relation prediction, it learns a predictive distribution by building a lightweight GCN-based, self-supervised entity typing model. 

Rosso et al. \cite{rosso2021reta} propose the RETA-Filter in the setting of instance completion, the task to retrieve relevant $(r,t)$-pairs for a given head entity. Like Borrego et al. \cite{borrego2019generating} and us (see Section \ref{sec:estimating}), they balance the trade-off between candidate coverage and search space reduction. In our setting, the RETA-filter suffers from scalability issues for a few reasons. First, the RETA-filter depends on creating a 3-dimensional matrix that becomes heavy to compute in large-scale settings, relying on computing the mode-n tensor product. Secondly, in an evaluation setting, the number of samples needed to be sampled grows at a rate of $\mathcal{O}(|\mathcal{E}|)$ as the RETA-filter considers the head, becoming significantly heavier in large-scale settings. Finally, the RETA-filter relies on knowing the types of entites, which often can be noisy or not be available.

\section{Relation recommendation}

\paragraph{Preliminaries} First, we define $\mathcal{R}$ as the set of relations and $\mathcal{E}$ as the set of entities. Furthermore, let $\mathcal{RS}_r$ denotes the set of tail entities, i.e., the domain, for a relation $r \in \mathcal{R}$ and $\mathcal{D}_r$ denotes the set of head entities, i.e., the domain, for a relation $r \in \mathcal{R}$. 

We define relational recommenders to be methods that build scores or probabilities for an entity of being a head or tail of a relation as they only rely on sampling based on the relation in a query. They derive a score matrix $\mathbf{X} \in \mathbb{R}^{|\mathcal{E}| \times |\mathcal{R}|}$, with a higher score indicating a higher likelihood of being an entity partaking in a domain or range. Here, $\mathcal{E}$ represents the set of entities and $\mathcal{R}$ the set of relations. One of the earlier works doing this is the work by Abedjan et al. \cite{abedjan2014amending}, using Association Rule Mining (ARM) \cite{agrawal1994fast}, later implemented as WD as discussed in Section \ref{sec:relatedwork}. In the next section, we present two adaptations of previous methods which we use in our experiments. We wish to emphasize that our point is not to create novel relational recommenders that outperform all other relational recommenders in all settings. Rather, our point is to show that these can be used as a tool. Depending on the circumstances, different relational recommenders can be used depending on, for example, whether high-quality entity types are available or not. In Table \ref{table:criteria}, we list different desirable properties of relational recommenders, indicating in which situations which ones might be useful. %

\subsection{Linear WD (L-WD)}

\begin{figure*}
    \centering
    \includegraphics[width=\linewidth]{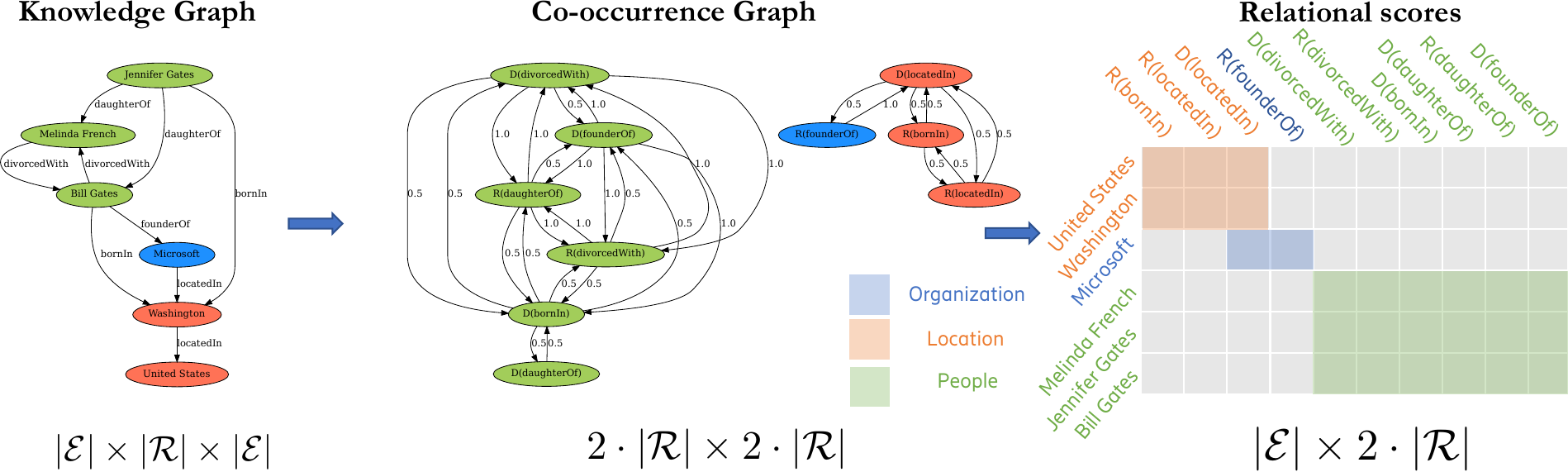}
    \caption{L-WD forms a global graph of confidence scores which are then aggregated into scores. These scores are later used to guide the selection of credible entities. In the co-occurrence graph, $D(\cdot)$-nodes represent domains of relations, $R(\cdot)$-nodes represent ranges.}
    \label{fig:L-WD}
\end{figure*}

L-WD is seen in Algorithm \ref{alg:L-WD} and illustrated in Figure \ref{fig:L-WD} on a toy example from Youn et al. \cite{youn-tagkopoulos-2023-kglm}. Given a KG $\mathcal{KG}_{train} \subset \mathcal{E} \times \mathcal{R} \times \mathcal{E}$, we extract the domains \& ranges as a binary matrix $\mathbf{B} \in \mathbb{R}^{|\mathcal{E}| \times 2 \cdot |\mathcal{R}|}, \mathbf{B}_{i,j} \in \{0,1\} \forall i,j$ where a 1 indicates that an entity has been seen as a head or tail of a relation. We then calculate a matrix $\mathbf{X} \in \mathbb{R}^{|\mathcal{E}| \times 2 \cdot |\mathcal{R}|}$ of the sum of aggregated ARM confidence scores \cite{agrawal1994fast} for each entity being a head or a tail of a relation. It therefore involves only two sparse matrix multiplications and a matrix normalization for any KG. The method is simple to compute, and does not require any information apart from the graph's structure ($\mathcal{KG}_{Train}$). Like WD \cite{abedjan2014amending}, L-WD can also be extended to incorporate types by treating them similarly to domains and ranges. We name this extension here \textbf{L-WD-T}. 

Intuitively, the matrix of confidence scores forms an adjacency matrix for a global graph of how domains \& ranges co-occur with each other. As an example, consider the relation \verb|ParentOf|, a relation naturally going from a person to a person. Equally, the relation \verb|LivesIn| goes naturally from a person to a location. This indicates that both the domain and range of \verb|ParentOf| and the domain of \verb|LivesIn| should co-occur frequently together, and something one can also observe in the graph. This co-occurrence adjacency matrix, $\mathbf{W}$ is formed by performing the matrix multiplication $\mathbf{B}^T\mathbf{B}$ and then normalizing $\mathbf{W}$ row-wise, generating probabilities between 0 and 1. The matrices used, $\mathbf{B}, \mathbf{W}$ and the resulting score matrix are generally very sparse, allowing for cheap operations and can therefore be run in seconds on a CPU.  Unlike WD, we do not use the average of the squared confidence scores and do not use a minimum confidence threshold. L-WD can therefore be considered a parameter-free version of rule mining and does not require any confidence or other hyperparameters or running multiple iterations. Instead, one only aggregates the scores through a series of sparse matrix multiplications. This gives L-WD desirable properties to be a seamless, easy-to-use candidate generator (see Table \ref{table:criteria}).

\begin{algorithm}
\caption{L-WD and L-WD-T}\label{alg:L-WD}
\begin{algorithmic}
\Require $\mathcal{KG} = \{(e_h, r, e_t) \subset \mathcal{E} \times \mathcal{R} \times \mathcal{E}\}$, Method $\in \{\text{L-WD}, \text{L-WD-T}\}$, (optional) typeset $\mathcal{TS} = \{(e, t) \subset \mathcal{E} \times \mathcal{T}\}$
\State Set Domains $\mathcal{D} = \{(e_h, r) \quad \forall (e_h, r, e_t) \in \mathcal{KG}\}$
\State Set Ranges $\mathcal{RS} = \{(e_t, r + |\mathcal{R}|) \quad \forall (e_h, r, e_t) \in \mathcal{KG}\}$
\State Set $\mathcal{DR} = \mathcal{D}\cup\mathcal{RS}$
\If {$\mathcal{TS} \neq \emptyset$}
    \State $\mathbf{B} \in \mathbf{R}^{|\mathcal{E}| \times 2 \cdot |\mathcal{R}|}, \mathbf{B}_{i,j} \in \{0,1\}$
\Else
    \State $\mathbf{B} \in \mathbf{R}^{|\mathcal{E}| \times 2 \cdot |\mathcal{R}| + |\mathcal{T}|}, \mathbf{B}_{i,j} \in \{0,1\}$
    \State Set $\mathcal{DR} = \mathcal{DR} \cup \mathcal{TS}$
\EndIf

\State Set $\mathbf{B}_{i,j} = \mathds{1}((e_i, x_j) \in \mathcal{DR}) \quad \forall i \in \{1,..,|\mathcal{E}|\}, j \in \{1, ..., 2\cdot |\mathcal{R}|\}$

\State Set $\mathbf{W} = \mathbf{B}^T\mathbf{B}$
\State Normalize $\mathbf{W}$ row-wise
\State Calculate score matrix $\mathbf{X} = \mathbf{B}\mathbf{W}$
\State Return score matrix $\mathbf{X}$
\end{algorithmic}
\end{algorithm}

\begin{table}[ht]
\centering
\caption{Desirable criteria for candidate generation methods. }
\label{table:criteria}
\resizebox{\linewidth}{!}{%
\begin{tblr}{
  column{even} = {c},
  column{1} = {r},
  column{3} = {c},
  column{5} = {c},
  hline{1,7} = {-}{0.08em},
  hline{2} = {1-3,6}{0.03em},
  hline{2} = {4-5}{},
}
                                          & \textbf{DBH} & \textbf{DBH-T} & \textbf{PIE} & \textbf{\textbf{L-WD-T }} & \textbf{L-WD } \\
\textbf{Scalable on CPU}                   & \textcolor{green}{\ding{52}}           & \textcolor{green}{\ding{52}}             & \textcolor{red}{\ding{56}}           & \textcolor{green}{\ding{52}}                              &
\textcolor{green}{\ding{52}}                   \\
\textbf{Parameter-free}                   & \textcolor{green}{\ding{52}}           & \textcolor{green}{\ding{52}}             & \textcolor{red}{\ding{56}}           & \textcolor{green}{\ding{52}}                              & \textcolor{green}{\ding{52}}                   \\
\textbf{Supports Unseen Candidates}       & \textcolor{red}{\ding{56}}           & \textcolor{green}{\ding{52}}             & \textcolor{green}{\ding{52}}           & \textcolor{green}{\ding{52}}                              & \textcolor{green}{\ding{52}}                   \\
\textbf{Type-free} & \textcolor{green}{\ding{52}}           & \textcolor{red}{\ding{56}}             & \textcolor{green}{\ding{52}}           & \textcolor{red}{\ding{56}}                              & \textcolor{green}{\ding{52}}                   \\
\textbf{Inductive}                        & \textcolor{red}{\ding{56}}           & \textcolor{green}{\ding{52}}             & \textcolor{green}{\ding{52}}           & \textcolor{green}{\ding{52}}                              & \textcolor{green}{\ding{52}}                   
\end{tblr}%
}
\end{table}

\subsection{DBH-T and OntoSim}

As DBH \cite{chensolution} is upper-bounded by PT, we create a stronger heuristic that combines entity types and domains \& ranges. If an entity $e \in \mathcal{E}$ is of type $t \in \mathcal{T}$ and is seen as a head for relation $r \in \mathcal{R}$, all entities of type $t$ will receive an added score of 1 for the domain of $r$. This heuristic shares characteristics with the RETA-Filter \cite{rosso2021reta} as both are based on counting the number of times a certain relation co-occurs with an entity of a certain type. However as DBH-T does not consider the head for query, it has a significantly lower sampling complexity for the evaluation than the RETA-filter, allowing orders of magnitude fewer samples during an evaluation, as we discuss in Section \ref{section:motivation}. 

We also consider the heuristic to assign all entities of a type $type$ to belong to a domain or range if any entity of type $t \in \mathcal{T}$ has been seen as a head or tail of the relation, calling it \textbf{OntoSim}.

\section{Motivation} \label{section:motivation}

In this section, we motivate our framework by analyzing a series of datasets and theoretically motivate why our framework using Relation Recommenders brings better estimations of the true rank.

\paragraph{The large amount of (extremely) easy negatives} \label{sec:easy-negatives}
\begin{table}[ht]
    \caption{Results from mining easy negatives is with L-WD. }
    \label{tab:falsenegs}
    \centering
    \resizebox{\linewidth}{!}{%
    \begin{tabular}{rrrr}
    \hline
    \textbf{}                             & \textbf{FB15k237} & \textbf{YAGO310} & \textbf{ogbl-wikikg2} \\ \hline
    \textbf{Easy negatives (\%)}                   & 58.4              & 43.2             & 5.42                  \\
    \textbf{Easy negatives}                        & 4,016,778         & 3,936,590        & 144,216,574           \\
    \textbf{False easy negatives} & 4                 & 0                & 35                    \\ \hline
    \end{tabular}%
    }
    
\end{table}

When investigating the scores of L-WD, we find that a significant amount of candidates can be quickly and confidently ruled out in many benchmark datasets through mining easy negatives. As mentioned, $\mathbf{X}$ are on many datasets very sparse, and if an entity has a score 0 for a certain domain, it can be considered highly unlikely it would ever be a head of the relation. On each dataset, only a handful out of the millions or even hundreds of millions of easy negatives mined constituted a part of a triple in the datasets (see Table \ref{tab:falsenegs}). Investigating these false negatives by hand, we find that several of these were semantically problematic due to different reasons. For example, we found the triple {(\sc MonthOfAugust}, {\sc /people/person/gender}, {\sc male}) to be included in the test set of FB15k237. We display all incorrectly classified examples in the Appendix, Section \ref{sec:false-negs-lWD}, Table \ref{table:falsehardnegs}. One can therefore not only directly rule out a large portion of negatives with a very high confidence by giving these a score of 0; it goes along the lines that one does not need to rank a large portion of the entities in the graph when predicting a head or tail and confidently derive an accurate estimate. The results of this support the conclusions of prior works \cite{safavi-koutra-2020-codex, wang-etal-2022-simkgc} --- while filtering out easy negatives in triplet classification is easy, classifying more difficult negatives is not. Precisely as in their work, we find that the more hard negative samples there are, the more difficult is the task. Therefore, when sampling randomly, one runs a high risk of sampling a negative that could have been ruled out almost instantly. Instead, we find more difficult negatives with Relational Recommenders.

\paragraph{Why sampling uniformly at random is inaccurate}

Here, we establish that a random sampling generally is an \textit{optimistic} estimation of the true ranking metric. The standard metrics commonly used in KGC are \textit{recall-based}, meaning that they rely only on the rank of each individual query being as high as possible. As the rank is a position of a true entity $t$ in a sorted list for a query (h, r, t), the ranking metrics depend purely on the position of $t$ in the list of sampled entities to rank. With this in mind, we can denote the number of (unknown) entities ranked higher than $t$ in this list to be $\mathcal{E}_{(h, r)}$. If we sample $n_s$ entities \textit{without replacement} for ranking, the number of entities landing in a higher position demoting the rank of the true answer follows a hypergeometric distribution $X_u \sim \text{H}(\lvert \mathcal{E}_{(h,r)}\rvert, \lvert \mathcal{E}\rvert, n_s)$. With $\lvert\mathcal{E}_{(h,r)}\rvert$ and $\lvert\mathcal{E}\rvert$ being constant, with the first only being known during a full evaluation and depending on our KGC model's predictive distribution, the variable to control for is $n_s$. It is trivial to prove Equation \ref{eq:lim}, but the implication is important; the smaller the sample size, the smaller the expected number of entities ranked higher than our true query when we evaluate the model's predicting capability. With Equation \ref{eq:lim} being true, the number of items worsening our ranking metrics decreases with $n_s$. Therefore, the smaller the sample sizes, the more optimistic will our expected ranking metrics be. Equally, as $n_s \rightarrow \lvert\mathcal{E}\rvert$, $\mathbb{E}[X_u] = \lvert \mathcal{E}_{(h,r)}\rvert$, meaning that the larger the sample, the closer to the true rank of our query we come.

\begin{equation} \label{eq:lim}
    \lim_{n_s \rightarrow 0} \mathbb{E}[X_u] = \lim_{n_s \rightarrow 0} n_s\frac{\lvert \mathcal{E}_{(h,r)}\rvert}{\lvert\mathcal{E}\rvert} = 0
\end{equation}

If we instead only sample relevant entities from, i.e., the tails set $\mathcal{RS}_r$, we will, on average, perform equally or better for any query $(h, r, ?)$ or $(?, r, t)$ as stated in Theorem \ref{theorem:guaranteed}. The proof is found in the Appendix, Section \ref{proof:theorem2}. 

\begin{theorem} \label{theorem:guaranteed}
    Let $\mathcal{KG} \subset \mathcal{E} \times \mathcal{R} \times \mathcal{E}$. For any query $(h, r, ?)$, let $t$ be the true answer and $\mathcal{E}_{(h,r)}$ be all entities having a higher rank than $t$ in a full evaluation. Let $Y$ be the number of positions closer to the true rank when sampling over the range set $\mathcal{RS}_r$ compared to the full entity set $\mathcal{E}$ and $0 < n_s \leq \lvert\mathcal{E}\rvert$. When sampling $n_s$ items to rank, $\mathbb{E}[Y] \geq 0$.
    
\end{theorem}

It is important to note that this holds under the assumption of well-defined domains \& ranges, meaning that they accurately represent which entities can partake in which relations which naturally depends on the quality of the KG. Furthermore, this assumes a uniform sampling across all entities in the sampling pool. To make it further robust to this noise, we derive probabilities through Relation Recommenders, biasing the outcome towards more credible entities. As these models have been proven accurate by previous work \cite{zangerle2016empirical}, they are suitable for separating non-credible entities from credible entities. Deriving an analytical form and guarantees is, in this case, very difficult, as it does not resemble a hypergeometric distribution as the probability of sampling an entity varies. Instead, we investigate the benefits of this empirically in our experiments. Furthermore, we also assume a KG containing entities of different types, with relations inherently connected to the types of entities. In other words, we do not consider KGs such as WordNet \cite{miller1995wordnet} and other KGs with a small set of relationship types and all entities are of the same type. 

\paragraph{Sampling efficiency} As the complexity of the evaluation is $\mathcal{O}(|\mathcal{E}|^2)$ as noted by Bastos et al. \cite{bastos2023can}, sampling a subset of candidates proposes a nonlinear decrease in terms of time needed for the evaluation. Sampling can however be computationally expensive on a large scale; in fact, a large portion of the time spent during training is attributed to sampling \cite{Hajimoradlou2022StayPK, li2021efficient}. If one uses a candidate generator that takes both the entity and relation into account, one needs to sample a set of entities for each distinct (h,r)- and (r,t)-pair in the evaluation set, growing at the rate of $\mathcal{O} (2 \cdot f_s  \cdot |\mathcal{E}| \cdot |\mathcal{KG}_{test}|)$ samples where $f_s$ is the fraction of entities sampled. If one instead uses Relation Recommenders this is avoided as they are agnostic to the entire query. As they do not consider the head, one can sample head- and tail candidates for each relation instead of for each query. This allows us to only do $2 \cdot |\mathcal{R}|$ samplings, drastically reducing the worst-case sampling complexity from $\Omega ( f_s  \cdot |\mathcal{E}|  \cdot |\mathcal{KG}_{test}|)$ to $\Omega(f_s  \cdot |\mathcal{E}|  \cdot 2 \cdot |\mathcal{R}|)$, where $f_s$ is the fraction of entities to sample from the entities. %

To show this with the datasets used in this study, one can investigate the worst-case scenario for the test sets. For example purposes, we assume sampling a maximum of 2.5 \% of all entities for each ranking on the three largest datasets used in our study; CoDEx-M \cite{safavi-koutra-2020-codex}, YAGO3-10 and ogbl-wikikg2 \cite{hu2020open} (the remaining can be seen in the Appendix, Table \ref{table:sample-efficiency}). If one uses a candidate generator during the evaluation that takes the head entity, one would need to sample the same number of times as there are distinct (h,r)- (r,t)-pairs in the test set, resulting in the number of samples as in the upper part of Table \ref{table:sample-efficiency}. Now, consider instead using a relational recommender in the same table to the right. Here, one only needs as many negative samples as twice the amount of relations to test. In this case, the number of false candidates to be sampled in total for an evaluation is consistently reduced by at least one order of magnitude across all datasets. Note also that this is independent of the fraction sampled. Therefore, the sampling using a relational recommender provides a significantly less expensive alternative in terms of sampling. 

\begin{table}[]
\caption{Number of samples needed during an evaluation with a candidate generator taking the entity into
account (above) and a relational recommender (below) at a sampling rate of 2.5 \%. }
\label{table:sample-efficiency}
\resizebox{\linewidth}{!}{%
\begin{tabular}{@{}ccrrr@{}}
\toprule
\textbf{Sampling}                                                  & \textbf{Dataset}                       & \multicolumn{1}{c}{\textbf{YAGO3-10}} & \multicolumn{1}{c}{\textbf{CoDEx-L}} & \multicolumn{1}{c}{\textbf{ogbl-wikikg2}} \\ \midrule
\multicolumn{1}{c|}{\multirow{2}{*}{$(h,r,\cdot), (\cdot, r, t)$}} & \textbf{\# (h,r)- \& (r,t)-pairs}      & 8,528                                 & 37,041                               & 645,483                                   \\
\multicolumn{1}{c|}{}                                              & \textbf{\# Samples}                    & 26,254,087                            & 72,184,574                           & 40,352,434,293                            \\ \midrule
\multicolumn{1}{c|}{\multirow{2}{*}{$(\cdot,r,\cdot)$}}            & \textbf{$(\cdot, r, \cdot)$-instances} & 34                                    & 65                                   & 367                                       \\
\multicolumn{1}{c|}{}                                              & \textbf{\# Samples}                    & 418,686                               & 506,681                              & 91,772,166                                \\ \midrule
\multicolumn{2}{c}{\textbf{Sampling reduction}}                                                             & x62,71                                & x142,47                              & x439,7                                   \\ \bottomrule
\end{tabular}%
}
\end{table}

\subsection{Estimating the true ranking metrics} \label{sec:estimating}

This section explains the two approaches of sampling entities proposed: Static (\textbf{S}) and Probabilistic (\textbf{P}) sampling. For both methods, negative entities are only sampled once from every head- and tail set, therefore $2 \cdot |\mathcal{R}|$ times.

\paragraph{Static} In the Static sampling, we define \textit{narrow} domains \& ranges by discretizing the score matrix $\mathbf{X} \in \mathbb{R}^{|\mathcal{E}| \times 2 \cdot |\mathcal{R}|}$ based on thresholding, giving a binary matrix that forms the sets of entities that are heads (the domain) or tails (the range) of a relation from the nonzero entries in each column. From these sets formed, we sample uniformly at random. When defining these sets, there are two conflicting objectives to consider; the Candidate Recall (CR) and the Reduction Rate (RR), i.e., the number of filtered-out candidates, as introduced by \cite{borrego2019generating}. The more candidates that are filtered out, the higher the risk of missing some true candidates. We adopt these two metrics in our work and use these to optimize this trade-off by finding the smallest $l_2$-distance to the optimal point $(CR, RR) = (1,1)$. To discretize each column in $\mathbf{X}$, we optimize the trade-off between the \textit{reduced search space} and \textit{the number of correctly predicted domains}, i.e., the recall. For clarity, the search space for one head- or tail set is its size, i.e., number of entities in each set. For each head- or tail-set, we set a probability threshold $T_{dr} \in [0,1], dr \in \mathcal{DR}$. This adapts the filtering for every domain and range, as some entities may have more relationship types than others. For example, an entity of type \verb|Country| such as \verb|Q142| (labelled \textit{France}) will be part of significantly more domains \& ranges than more specific attributes (75 to be exact), such as the attribute \verb|Q14763015| (labelled \textit{protein-containing complex localization}) in ogbl-wikikg2, only being at the end of the relation \verb|P682| (labelled \textit{biological process}), therefore only participating in one domain/range. 

One might consider, if domains and ranges are available through an underlying ontology, to use these directly. However, an ontology might not always be available, and types are often incomplete and noisy \cite{yao2021typing,hu-etal-2022-transformer}. On our workloads, we simulate this in the experiments and finds this to risk becoming overly precise, not reducing the set sizes as much as desired. 

\paragraph{Probabilistic} In the Probabilistic setting, we use the score matrix $\mathbf{X}$ to sample candidates using the probabilities. This allows us to sample more credible entities. For example, if we wish to evaluate a queries $(h, \text{livesIn}, ?)$,  we sample from range column of the relation \verb|livesIn| in $\mathbf{X}$. A higher likelihood of an entity being a tail of this relation will then entail a higher chance of being sampled.

\section{Experiments}

Our experiments consist of three parts; Section \ref{section:comparing} describes how we compare different relation recommenders. Section \ref{section:comparing} describes our comparison of different estimators of ranking metrics and their correlation to the true metric, and Section \ref{section:large-scale-exp} explains our large-scale experiments on ogbl-wikikg2. The datasets we use in this study can be seen in Table \ref{table:datasets}.

\begin{table}[]
\caption{Statistics of the datasets used in this study.}
\label{table:datasets}
\resizebox{\linewidth}{!}{%
\begin{tabular}{@{}crrrrrrrrr@{}}
\toprule
\multicolumn{1}{l|}{\textbf{}}             & \multicolumn{4}{c}{}                          & \multicolumn{3}{c}{\textbf{Triples}}                                                             & \multicolumn{2}{c}{\textbf{(h,r) \& (r,t)-pairs}}                \\ \midrule
\textbf{Dataset}          & $|\mathcal{E}|$     & $|\mathcal{R}|$ & $|\mathcal{T}|$ & $|\mathcal{TS}|$ & \textbf{Train} & \textbf{Valid} & \textbf{Test} & \textbf{Train} & \textbf{Test} \\ \midrule
\multicolumn{1}{c|}{\textbf{FB15k}}    & 14,505    & 1,345             & 79              & 76,752           & 272,115                         & 20,438                        & 17,526                         & 149,689                         & 12,193                         \\
\multicolumn{1}{c|}{\textbf{FB15k-237}}    & 14,505    & 237             & 79              & 76,752           & 272,115                         & 20,438                        & 17,526                         & 149,689                         & 12,193                         \\
\multicolumn{1}{c|}{\textbf{YAGO3-10}}     & 123,143   & 37              & 32574           & 7,028,143        & 1,079,040                       & 4,982                         & 4,978                          & 392,122                         & 1,826                          \\
\multicolumn{1}{c|}{\textbf{ogbl-wikikg2}} & 2,500,604 & 535             & 9,322           & 2,558,406        & 16,109,182                      & 429,456                       & 598,543                        & 14,941,073                      & 859,527                        \\
\multicolumn{1}{c|}{\textbf{CoDEx-S}}      & 2034      & 42              & 2034               & 147                & 32,888                          & 1827                          & 1828                           & 11,867                          & 369                            \\
\multicolumn{1}{c|}{\textbf{CoDEx-M}}      & 17,050    & 51              & 17,050               & 483                & 185,584                         & 10310                         & 10311                          & 101,136                         & 3559                           \\
\multicolumn{1}{c|}{\textbf{CoDEx-L}}      & 77,951    & 69              & 77,951               & 1,405                & 551,193                         & 30,622                        & 30,622                         & 380,303                         & 14,778                         \\ \bottomrule
\end{tabular}%
}
\end{table}

\subsection{Comparing relational recommenders} \label{section:comparing}

First, we compare relation recommenders' ability to generate accurate yet narrow relational head- and tail sets. 

\paragraph{Metrics} As discussed in Section \ref{sec:estimating}, the Reduction Rate (RR) and the Candidate Recall (CR) exhibit a trade-off which we optimize. For Candidate Recall, we report two versions: \textbf{Test} and \textbf{Unseen}. \textbf{Test} refers to the metric measured on \textit{all} (h,r)- and (r,t)-pairs in the test set, while \textbf{Unseen} reflects the recall on the ones not seen in the training or validation set. For example, more common domain or range examples such as $\texttt{Male} \in R(\texttt{HasGender})$ have already been seen, and can be directly incorporated by simply observing the training set. Moreover, we include the already seen entities when measuring \textbf{CR Test}, therefore combining \textbf{PT} with each method to simulate a practical scenario where one naturally would do this.

\paragraph{How does the choice of relation recommender affect the estimation?} Although our intention is not to create a novel relational recommender, but rather utilize them in a framework, we consider various relation recommenders to learn the domains and ranges. First, from the OGB-LSC challenge \cite{hu2021ogblsc} we consider PIE. For our settings on training PIE, we refer to Appendix, Section \ref{sec:pie-training}. We also compare with PseudoTyped (PT) and the other heuristics discussed in \ref{sec:relatedwork}. For example, if an entity of type \verb|Person| has been seen as a head of \verb|LivesIn|, all other people will be included in the domain of \verb|LivesIn|. For all methods using types (L-WD-T, DBH-T and OntoSim), we use the same types for each method. How these are obtained for each dataset is described in the Appendix, Section \ref{sec:obtaining-entity-types}.

To further provide a comprehensive study on the impact of the choice of relational recommender in our framework, we investigate the MAPE compared to the sample sizes needed to accurately estimate the true performance. As the number of samples decrease, the estimate becomes more optimistic. On the test sets, we evaluate the set of fully trained models to determine which of the relational recommenders is the most efficient. For each relational recommender, we sample five times using both a probabilistic and a static approach. We then set the sample fraction over a range of values between 0.01 and 0.3.

\subsection{Measuring correlations and error rate} \label{section:corr-error-exps}

In our second part of our experiments, we train KGC models on a wide range of datasets and KGC models; we use TransE \cite{bordes2013translating}, ComplEx \cite{trouillon2016complex}, DistMult \cite{yang2014embedding}, ConvE \cite{dettmers2018conve}, TuckER \cite{balazevic2019tucker}, RESCAL \cite{nickel2011three}. For training on the smaller datasets, we use the \verb|LibKGE| package \cite{libkge} and leverage the configurations of the optimally tuned models\footnote{\url{https://github.com/uma-pi1/kge}}. We run for 100 epochs and evaluate the full ranking metrics on every validation epoch and estimate them using the different methods.

We compare our proposed sampling techniques with $\mathcal{KP}$ \cite{bastos2023can} and rank estimations when sampling uniformly at random (abbreviated \textbf{R}) and $\mathcal{KP}$ whose code we adopt\footnote{\url{https://github.com/ansonb/Knowledge_Persistence}}. We also investigate whether our method of sampling negatives can help $\mathcal{KP}$ in improving the correlation with the ranking metric and report $\mathcal{KP}$ with random uniform sampling (\textbf{R}), probabilistic sampling, \textbf{P} and static sampling (\textbf{S}).

We report two metrics: the Pearson correlation and the Mean Absolute Error (MAE) from the true, full-ranking metric for the instances intended to predict it. A high correlation with the full, true ranking metric is desirable as it indicates whether the quicker method can replace the existing evaluation, but the MAE allows us to determine whether it correctly predicts the \textbf{true} metric. For these experiments, we use L-WD to guide our sampling as it overall generalizes to most settings, showing that the framework generalizes to a range of settings. We set the sampling to be $10 \%$ of all entities, i.e., $n_s = 0.1 \cdot |\mathcal{E}|$ for all sampling methods for all datasets, except for ogbl-wikikg2 where we set $n_s = 200,000$ $(\approx 0.08)$.

A crucial part when determining a method's capability of estimating the true performance is to see how different methods rank the performance of models in compared to their rank in terms of the true metric. To do this, we measure the Kendall-Tau rank correlations \cite{abdi2007kendall} on the datasets where we have trained three or more models. During each epoch, we measure how the model ranking of each performance estimation correlates with the ranking of the models by the true performance. This allows us to measure whether one would, during each epoch, be able to determine the highest performing model.

\subsection{In-depth investigation on ogbl-wikikg2} \label{section:large-scale-exp}

On ogbl-wikikg2 we evaluate during training, but also take pretrained Complex embeddings\footnote{\url{https://github.com/facebookresearch/ssl-relation-prediction/}} from Chen et al. \cite{chen2021relation} and evaluate on the test set. While on the smaller datasets, even the full evaluation runs in seconds on most datasets when we use LibKGE, a full evaluation takes substantially longer to run due to the quadratic increase in computational complexity. We measure the time taken to evaluate for a range of number of samples, and compare how many samples $n_s$ are needed to correctly estimate the true ranking metrics and what its time consumption is. %

\begin{table}[ht]
\caption{Results on Candidate Recall (CR), Reduction Rate (RR), and Reduced Scoring (RS) on the test set. Best numbers in bold, runner-up(s) underlined. }
\label{table:candidate-recall-reltypes}
\resizebox{\linewidth}{!}{%
\begin{tabular}{crrrc}
\hline
\textbf{Dataset}                       & \multicolumn{1}{c}{\textbf{Model}}        & \multicolumn{1}{c}{\textbf{CR (Test/Unseen)}} & \multicolumn{1}{c}{\textbf{RR}} & \textbf{Runtime} \\ \hline
\multirow{6}{*}{\textbf{FB15k237}}     & \multicolumn{1}{r|}{\textbf{PT}} & 0.838/0                                       & \textbf{0.977}                  & 0.09 sec         \\
                                       & \multicolumn{1}{r|}{\textbf{DBH-T}}       & 0.978/0.864                                   & 0.875                           & 2.41 sec         \\
                                       & \multicolumn{1}{r|}{\textbf{OntoSim}}       & \textbf{0.998/0.987}                                   & 0.416                           & 2.27 sec         \\
                                       & \multicolumn{1}{r|}{\textbf{PIE}}         & {0.982/0.889}                             & 0.928                           & 210 min          \\
                                       & \multicolumn{1}{r|}{\textbf{L-WD}}   & {0.982/0.889}                             & 0.932                           & 0.22 sec         \\
                                       & \multicolumn{1}{r|}{\textbf{L-WD-T}} & {\ul 0.984/0.902}                          & {\ul 0.937}                     & 4.66 sec         \\ \hline
\multirow{6}{*}{\textbf{YAGO310}}      & \multicolumn{1}{r|}{\textbf{PT}} & 0.907/0                                       & \textbf{0.957}                  & 0.29 sec         \\
                                       & \multicolumn{1}{r|}{\textbf{DBH-T}}       & 0.986/0.849                                   & 0.870                           & 80.82 sec        \\
                                       & \multicolumn{1}{r|}{\textbf{OntoSim}}       & \textbf{0.994/0.944}                                   & 0.113                           & 73.54 sec        \\
                                       & \multicolumn{1}{r|}{\textbf{PIE}}         & 0.981/0.799                                   & 0.895                           & 242 min          \\
                                       & \multicolumn{1}{r|}{\textbf{L-WD}}   & 0.985/0.835                             & 0.900                           & 0.43 sec         \\
                                       & \multicolumn{1}{r|}{\textbf{L-WD-T}} & {\ul 0.988/0.866}                          & {\ul 0.917}                     & 81.18 sec        \\ \hline
\multirow{6}{*}{\textbf{ogbl-wikikg2}} & \multicolumn{1}{r|}{\textbf{PT}} & 0.520/0                                       & \textbf{0.994}                  & 5.58 sec         \\
                                       & \multicolumn{1}{r|}{\textbf{DBH-T}}       & 0.598/0.163                                   & 0.954                           &               300.6 sec   \\
                                       & \multicolumn{1}{r|}{\textbf{OntoSim}}       &         \textbf{0.982/0.962}                           &   0.75                         &               334.68 sec   \\
                                       & \multicolumn{1}{r|}{\textbf{PIE}}         & 0.896/0.783                                   & 0.936                           & 2 days           \\
                                       & \multicolumn{1}{r|}{\textbf{L-WD}}   & {\ul 0.910/0.812}                             & 0.939                           & 16.53 sec        \\
                                       & \multicolumn{1}{r|}{\textbf{L-WD-T}} & {\ul 0.912/0.817}                          & {\ul 0.951}                     &  389.3 sec                \\ \hline
\end{tabular}%
}
\end{table}
\section{Results and Discussion} \label{sec:results}

\begin{table}[]
\caption{MAEs of estimating the Filtered validation MRR with different sampling strategies. Bold indicates best estimator. }
\label{table:MAE-estimators}
\begin{tabular}{@{}crrrr@{}}
\toprule
\textbf{Dataset}                    & \multicolumn{1}{c}{\textbf{Model}} & \multicolumn{1}{c}{\textbf{R}} & \multicolumn{1}{c}{\textbf{P}} & \multicolumn{1}{c}{\textbf{S}} \\ \midrule
\multirow{7}{*}{\textbf{FB15k-237}} & \textbf{TransE}                    & 0.216                          & 0.017                          & \textbf{0.008}                 \\
                                    & \textbf{RotatE}                    & 0.228                          & 0.017                          & \textbf{0.006}                 \\
                                    & \textbf{RESCAL}                    & 0.225                          & 0.016                          & \textbf{0.002}                 \\
                                    & \textbf{DistMult}                  & 0.224                          & 0.016                          & \textbf{0.005}                 \\
                                    & \textbf{ConvE}                     & 0.213                          & 0.010                          & \textbf{0.001}                 \\
                                    & \textbf{ComplEx}                   & 0.226                          & 0.015                          & \textbf{0.003}                 \\ \midrule
\multirow{6}{*}{\textbf{FB15k}}     & \textbf{TransE}                    & 0.212                          & 0.042                          & \textbf{0.014}                 \\
                                    & \textbf{RotatE}                    & 0.151                          & 0.031                          & \textbf{0.012}                 \\
                                    & \textbf{RESCAL}                    & 0.214                          & 0.026                          & \textbf{0.003}                 \\
                                    & \textbf{DistMult}                  & 0.102                          & 0.014                          & \textbf{0.004}                 \\
                                    & \textbf{ConvE}                     & 0.187                          & 0.020                          & \textbf{0.006}                 \\
                                    & \textbf{ComplEx}                   & 0.074                          & 0.012                          & \textbf{0.004}                 \\ \midrule
\multirow{4}{*}{\textbf{CoDEx-S}}   & \textbf{TransE}                    & 0.307                          & 0.071                          & \textbf{0.008}                 \\
                                    & \textbf{RESCAL}                    & 0.273                          & 0.069                          & \textbf{0.002}                 \\
                                    & \textbf{ConvE}                     & 0.262                          & 0.061                          & \textbf{0.001}                 \\
                                    & \textbf{ComplEx}                   & 0.249                          & 0.061                          & \textbf{0.005}                 \\ \midrule
\multirow{2}{*}{\textbf{CoDEx-M}}   & \textbf{ConvE}                     & 0.183                          & 0.022                          & \textbf{0.001}                 \\
                                    & \textbf{ComplEx}                   & 0.179                          & 0.054                          & \textbf{0.007}                 \\ \midrule
\multirow{5}{*}{\textbf{CoDEx-L}}   & \textbf{TransE}                    & 0.185                          & 0.062                          & \textbf{0.049}                 \\
                                    & \textbf{TuckER}         & 0.154                          & 0.027                          & \textbf{0.002}                 \\
                                    & \textbf{RESCAL}                    & 0.150                          & 0.024                          & \textbf{0.002}                 \\
                                    & \textbf{ConvE}                     & 0.147                          & 0.021                          & \textbf{0.002}                 \\
                                    & \textbf{ComplEx}                   & 0.137                          & 0.022                          & \textbf{0.004}                 \\ \midrule
\textbf{YAGO3-10}                   & \textbf{ComplEx}                   & 0.187                          & 0.100                          & \textbf{0.030}                 \\ \midrule
\textbf{ogbl-wikikg2}                  & \textbf{ComplEx-RP}                   & 0.094                 & \textbf{0.002}                 & 0.012                \\
\bottomrule
\end{tabular}
\end{table}

\begin{table}[]
\caption{Correlation with the Filtered MRR. Bold indicates best, underlined the best sampling for $\mathcal{KP}$. }
\label{table:correlations}
\resizebox{\linewidth}{!}{%
\begin{tabular}{@{}crlll|lll@{}}
\toprule
\multicolumn{1}{l}{}                & \multicolumn{1}{l}{}                   & \multicolumn{3}{c|}{$\mathcal{KP}$}                                                                  & \multicolumn{3}{c}{\textbf{Rank estimates}}                                                            \\ \midrule
\textbf{Dataset}                    & \multicolumn{1}{c}{\textbf{Model}}     & \multicolumn{1}{c}{\textbf{R}} & \multicolumn{1}{c}{\textbf{P}} & \multicolumn{1}{c|}{\textbf{S}} & \multicolumn{1}{c}{\textbf{R}} & \multicolumn{1}{c}{\textbf{P}} & \multicolumn{1}{c}{\textbf{S}} \\ \midrule
\multirow{7}{*}{\textbf{FB15k-237}} & \multicolumn{1}{r|}{\textbf{TransE}}   & {\ul -0.235}                   & -0.479                         & -0.448                          & \textbf{0.997}                 & 0.984                          & 0.975                          \\
                                    & \multicolumn{1}{r|}{\textbf{RotatE}}   & {\ul 0.514}                    & 0.069                          & -0.096                          & \textbf{0.997}                 & 0.994                          & 0.983                          \\
                                    & \multicolumn{1}{r|}{\textbf{RESCAL}}   & 0.062                          & {\ul 0.791}                    & 0.742                           & \textbf{0.996}                 & 0.995                          & 0.994                          \\
                                    & \multicolumn{1}{r|}{\textbf{DistMult}} & {\ul 0.908}                    & 0.839                          & 0.852                           & \textbf{0.997}                 & 0.993                          & 0.994                          \\
                                    & \multicolumn{1}{r|}{\textbf{ConvE}}    & 0.516                          & {\ul 0.803}                    & 0.755                           & 0.993                          & \textbf{0.999}                 & \textbf{0.999}                 \\
                                    & \multicolumn{1}{r|}{\textbf{ComplEx}}  & {\ul 0.897}                    & 0.838                          & 0.847                           & 0.994                          & \textbf{0.997}                 & 0.991                          \\ \midrule
\multirow{6}{*}{\textbf{FB15k}}     & \multicolumn{1}{r|}{\textbf{TransE}}   & 0.029                          & {\ul 0.100}                    & 0.028                           & 0.992                          & \textbf{1.000}                 & \textbf{1.000}                 \\
                                    & \multicolumn{1}{r|}{\textbf{RotatE}}   & -0.068                         & {\ul 0.301}                    & 0.154                           & 0.974                          & \textbf{1.000}                 & 0.999                          \\
                                    & \multicolumn{1}{r|}{\textbf{RESCAL}}   & {\ul -0.140}                   & -0.389                         & -0.346                          & 0.993                          & \textbf{1.000}                 & \textbf{1.000}                 \\
                                    & \multicolumn{1}{r|}{\textbf{DistMult}} & 0.120                          & {\ul 0.589}                    & 0.548                           & 0.974                          & \textbf{1.000}                 & \textbf{1.000}                 \\
                                    & \multicolumn{1}{r|}{\textbf{ConvE}}    & 0.893                          & {\ul 0.940}                    & 0.929                           & 0.987                          & \textbf{1.000}                 & 0.999                          \\
                                    & \multicolumn{1}{r|}{\textbf{ComplEx}}  & 0.083                          & 0.177                          & {\ul 0.194}                     & 0.984                          & \textbf{1.000}                 & \textbf{1.000}                 \\ \midrule
\multirow{4}{*}{\textbf{CoDEx-S}}   & \multicolumn{1}{r|}{\textbf{TransE}}   & 0.142                          & {\ul 0.605}                    & 0.558                           & 0.944                          & 0.988                          & \textbf{0.999}                 \\
                                    & \multicolumn{1}{r|}{\textbf{RESCAL}}   & -0.182                         & {\ul 0.124}                    & 0.036                           & 0.856                          & 0.986                          & \textbf{1.000}                 \\
                                    & \multicolumn{1}{r|}{\textbf{ConvE}}    & -0.183                         & {\ul 0.434}                    & 0.286                           & 0.947                          & 0.996                          & \textbf{1.000}                 \\
                                    & \multicolumn{1}{r|}{\textbf{ComplEx}}  & 0.702                          & 0.889                          & {\ul 0.909}                     & 0.982                          & 0.995                          & \textbf{0.997}                 \\ \midrule
\multirow{2}{*}{\textbf{CoDEx-M}}   & \multicolumn{1}{r|}{\textbf{ConvE}}    & {\ul 0.773}                    & 0.761                          & 0.745                           & 0.971                          & 0.999                          & \textbf{1.000}                 \\
                                    & \multicolumn{1}{r|}{\textbf{ComplEx}}  & 0.403                          & 0.684                          & {\ul 0.780}                     & 0.967                          & 0.998                          & 1.000                          \\ \midrule
\multirow{5}{*}{\textbf{CoDEx-L}}   & \multicolumn{1}{r|}{\textbf{TransE}}   & 0.018                          & 0.065                          & {\ul 0.161}                     & \textbf{0.786}                 & 0.730                          & 0.745                          \\
                                    & \multicolumn{1}{r|}{\textbf{TuckER}}   & -0.180                         & {\ul 0.262}                    & 0.188                           & 0.882                          & 0.997                          & \textbf{1.000}                 \\
                                    & \multicolumn{1}{r|}{\textbf{RESCAL}}   & {\ul 0.345}                    & 0.245                          & 0.273                           & 0.919                          & 0.998                          & \textbf{1.000}                 \\
                                    & \multicolumn{1}{r|}{\textbf{ConvE}}    & {\ul 0.871}                    & 0.852                          & 0.851                           & 0.980                          & 0.999                          & \textbf{1.000}                 \\
                                    & \multicolumn{1}{r|}{\textbf{ComplEx}}  & -0.166                         & 0.049                          & 0.125                           & 0.388                          & \textbf{0.996}                 & 0.991                          \\ \midrule
\textbf{YAGO3-10}                   & \multicolumn{1}{r|}{\textbf{ComplEx}}  & 0.527                          & 0.632                          & {\ul 0.679}                     & 0.980                          & 0.958                          & \textbf{0.985}                 \\ \midrule
\textbf{ogbl-wikikg2}                   & \multicolumn{1}{r|}{\textbf{ComplEx-RP}}  & 0.956                          & ---                          & ---                     & \textbf{1.000}                          & 0.999                          & \textbf{0.985}                 \\\bottomrule
\end{tabular}%
}
\end{table}

\begin{table}[]
\caption{Average Kendall-Tau rank correlations of ranking models' performance over 100 epochs. }
\label{table:kendall-tau-correlations}
\begin{tabular}{@{}c|rrr|rrr@{}}
\toprule
\multicolumn{1}{l}{}     & \multicolumn{3}{c|}{$\mathcal{KP}$}                                                               & \multicolumn{3}{c}{\textbf{Rank estimates}}                                                      \\ \midrule
\textbf{Dataset}      & \multicolumn{1}{c}{\textbf{R}} & \multicolumn{1}{c}{\textbf{P}} & \multicolumn{1}{c|}{\textbf{S}} & \multicolumn{1}{c}{\textbf{R}} & \multicolumn{1}{c}{\textbf{P}} & \multicolumn{1}{c}{\textbf{S}} \\ \midrule
\textbf{FB15k-237}     & 0,018                          & -0,137                         & -0,106                          & 0,718                          & {\ul 0,876}                    & \textbf{0,909}                 \\
\textbf{FB15k}          & 0,064                          & 0,221                          & 0,194                           & 0,892                          & {\ul 0,968}                    & \textbf{0,972}                 \\
\textbf{CoDEx-S}        & 0,413                          & 0,703                          & 0,7                             & 0,663                          & {\ul 0,9}                      & \textbf{0,993}                 \\
\textbf{CoDEx-L}        & 0,372                          & 0,414                          & 0,41                            & 0,722                          & {\ul 0,85}                     & \textbf{0,908}                 \\ \bottomrule
\end{tabular}

\end{table}

\begin{table*}[]
\caption{Average speed-up of evaluation on the smaller datasets. Higher is better. }
\label{table:speed-up}
\resizebox{\linewidth}{!}{%
\begin{tabular}{@{}ccrrcrrrr@{}}
\toprule
\multicolumn{1}{l}{\textbf{}}             & \textbf{Sampling method}                                & \multicolumn{1}{c}{\textbf{CoDEx-S}} & \multicolumn{1}{c}{\textbf{CoDEx-M}} & \textbf{CoDEx-L}       & \multicolumn{1}{c}{\textbf{FB15k}} & \multicolumn{1}{c}{\textbf{FB15k-237}} & \multicolumn{1}{c}{\textbf{YAGO3-10}} & \multicolumn{1}{c}{\textbf{ogbl-wikikg2}} \\ \midrule
\multirow{3}{*}{\textbf{$\mathcal{KP}$}}            & \multicolumn{1}{c|}{\textbf{Random}}                    & \textbf{$8.7 \pm 2.0$}               & \textbf{$7.8 \pm 1.2$}               & $7.5 \pm 2.2$          & \textbf{$15.4 \pm 3.3$}            & \textbf{$8.5 \pm 2.8$}                 & \textbf{$4.4 \pm 0.3$}                & $18,66 \pm 0,633$                         \\
                                          & \multicolumn{1}{c|}{\textbf{Probabilistic}}             & $7.4 \pm 2.5$                        & $4.4 \pm 0.5$                        & $3.5 \pm 1.0$          & $12.4 \pm 2.8$                     & $6.2 \pm 2.1$                          & $0.9 \pm 0.2$                         & ---                                       \\
                                          & \multicolumn{1}{c|}{\textbf{Static}}                    & $7.9 \pm 1.9$                        & $4.4 \pm 0.5$                        & $3.5 \pm 1.0$          & $12.4 \pm 2.8$                     & $6.0 \pm 2.0$                          & $0.9 \pm 0.2$                         & ---                                       \\ \midrule
\multirow{3}{*}{\textbf{Ranking metrics}} & \multicolumn{1}{c|}{\textbf{Random}}                    & $2.7 \pm 0.3$                        & $5.9 \pm 0.5$                        & \textbf{$7.6 \pm 0.7$} & $2.8 \pm 0.4$                      & $3.6 \pm 1.1$                          & $2.3 \pm 0.1$                         & $56,3 \pm 6$                 \\
                                          & \multicolumn{1}{c|}{\textbf{Probabilistic}}             & $2.4 \pm 0.3$                        & $5.5 \pm 0.5$                        & $7.5 \pm 1.1$          & $2.4 \pm 0.4$                      & $3.2 \pm 1.0$                          & $1.9 \pm 0.2$                         & $115,9 \pm 3,5$                           \\
                                          & \multicolumn{1}{c|}{\textbf{Static}}                    & $2.4 \pm 0.3$                        & $5.2 \pm 0.4$                        & $5.3 \pm 0.5$          & $2.4 \pm 0.3$                      & $3.1 \pm 1.0$                          & $1.6 \pm 0.2$                         & \textbf{$141,8 \pm 4,4$}                  \\ \midrule
                                          \multicolumn{2}{c|}{\textbf{Full evaluation (seconds)}} & $0.7 \pm 0.1$                        & $4.2 \pm 0.4$                        & $15.6 \pm 3.8$         & $21.2 \pm 3.0$                     & $7.0 \pm 2.2$                          & $1.2 \pm 0.1$                         & $1566,4 \pm 26,4$                            \\ \bottomrule
\end{tabular}%
}
\end{table*}

\begin{figure*}[ht]

  \begin{subfigure}{0.33\textwidth}
  \caption{(Logarithmic) Evaluation time vs. sample size on the ogbl-wikikg2 test set. } \label{fig:1a-b}
    \includegraphics[width=\linewidth]{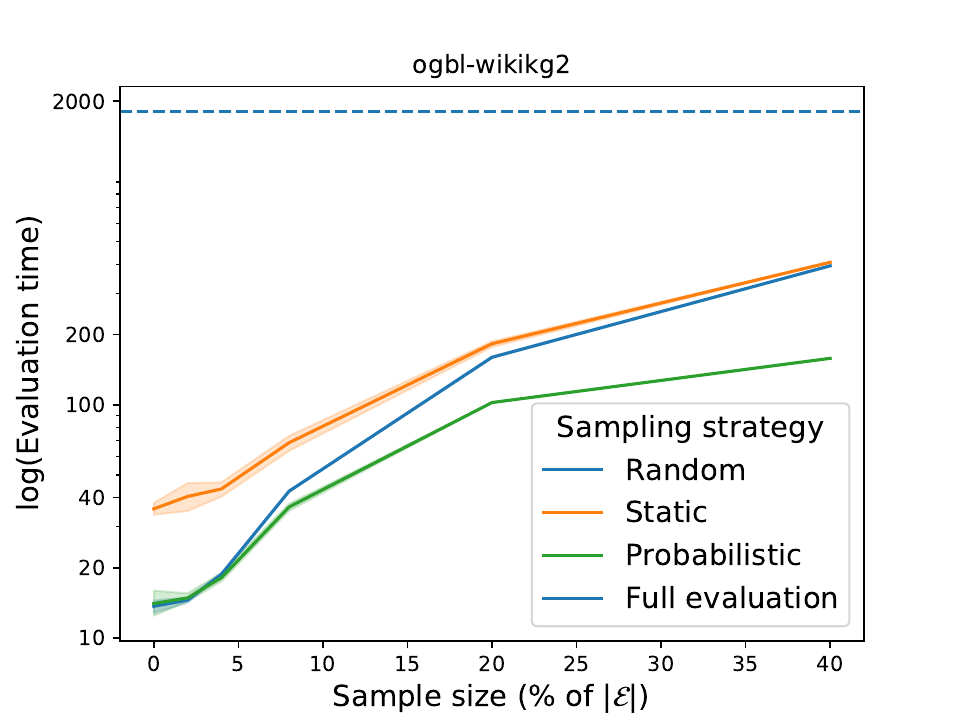}
    
  \end{subfigure}%
  \hspace*{\fill}   %
  \begin{subfigure}{0.33\textwidth}
  \caption{Filtered MRR vs. sample size on the ogbl-wikikg2 test set. } \label{fig:1b-b}
    \includegraphics[width=\linewidth]{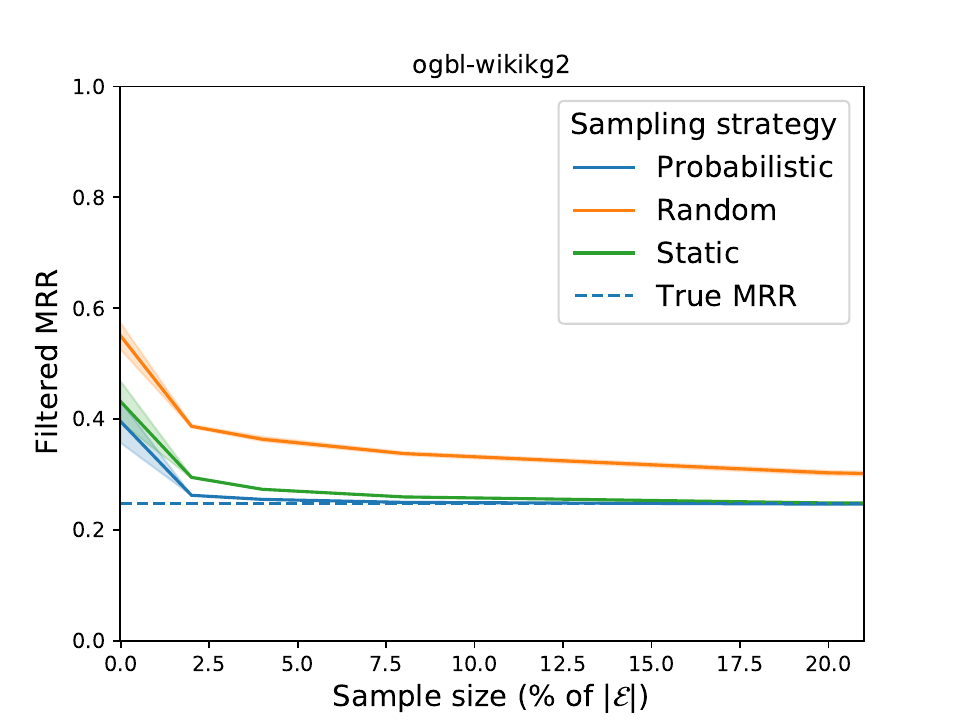}
    
  \end{subfigure}%
    \hspace*{\fill}   %
    \begin{subfigure}{0.33\textwidth}
    \caption{Estimated validation MRR on ogbl-wikikg2 across evaluations. }
    \includegraphics[width=\linewidth]{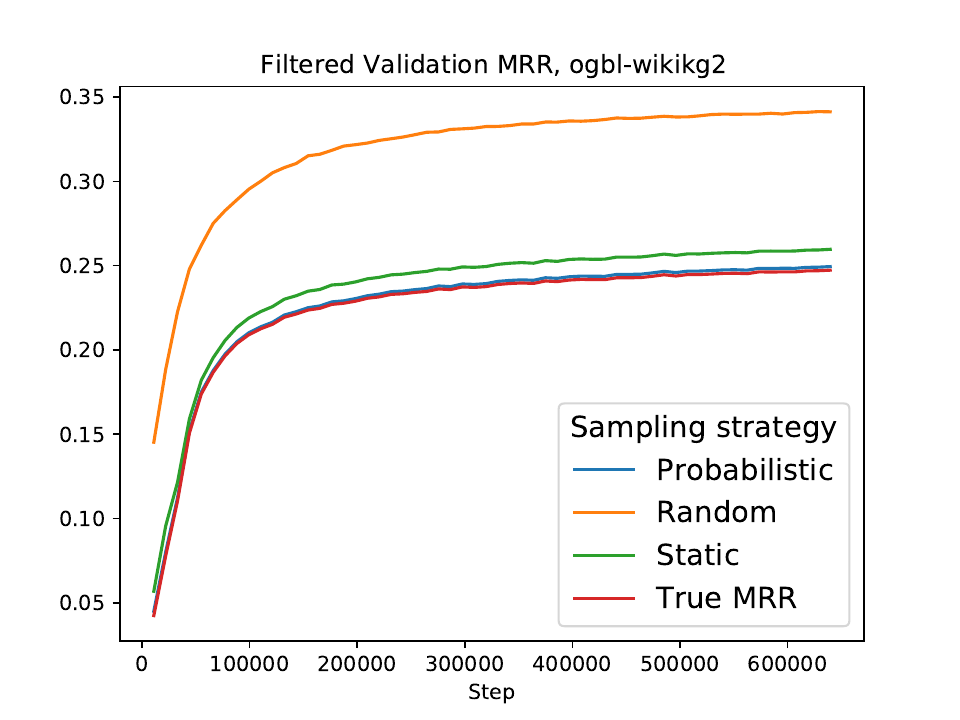}
    
   \label{fig:estimated-mrrs-ogbl-wikikg2}
    \end{subfigure}%
\caption{Left: a logarithmic scale of the evaluation time (seconds) against the sample size on ogbl-wikikg2. Full evaluation time is the dashed line. Middle: we display the Sample size (\% of $|\mathcal{E}|$) compared to Filtered MRR estimate of different sampling strategies, dashed line is true value. Right: The estimated validation MRRs of the different methods across training on ogbl-wikikg2. } \label{fig:1}
\end{figure*}

\begin{figure*}[ht]

  \begin{subfigure}{0.33\textwidth}
  \caption{ FB15k. } \label{fig:2a-b}
    \includegraphics[width=\linewidth]{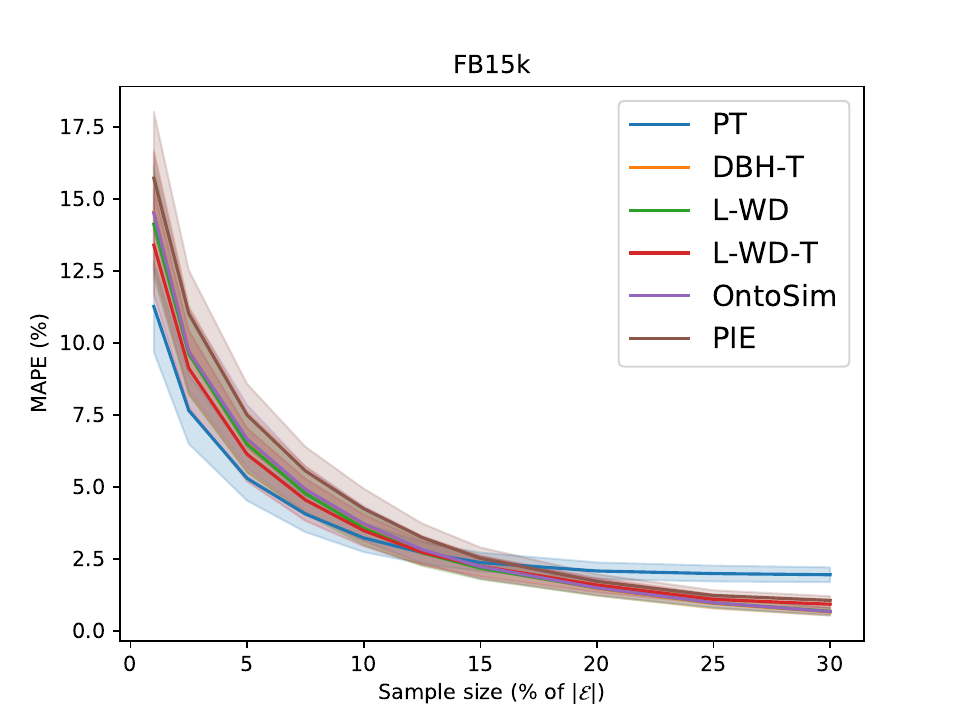}
    
  \end{subfigure}%
  \hspace*{\fill}   %
  \begin{subfigure}{0.33\textwidth}
  \caption{CoDEx-M } \label{fig:2b-b}
    \includegraphics[width=\linewidth]{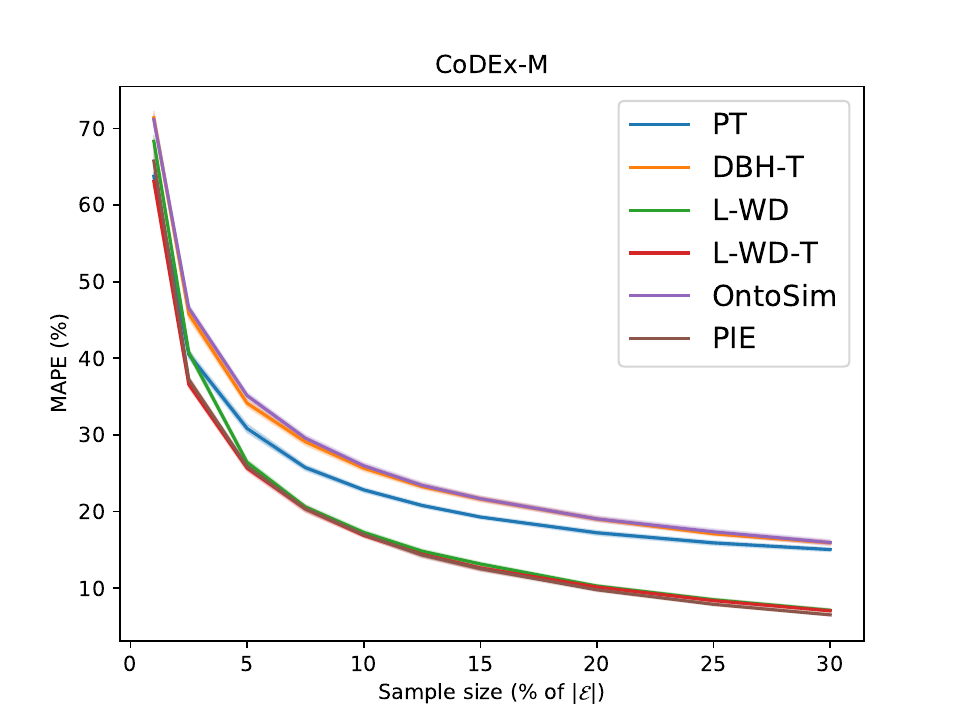}
    
  \end{subfigure}%
    \hspace*{\fill}   %
    \begin{subfigure}{0.33\textwidth}
    \caption{YAGO3-10.}
    \includegraphics[width=\linewidth]{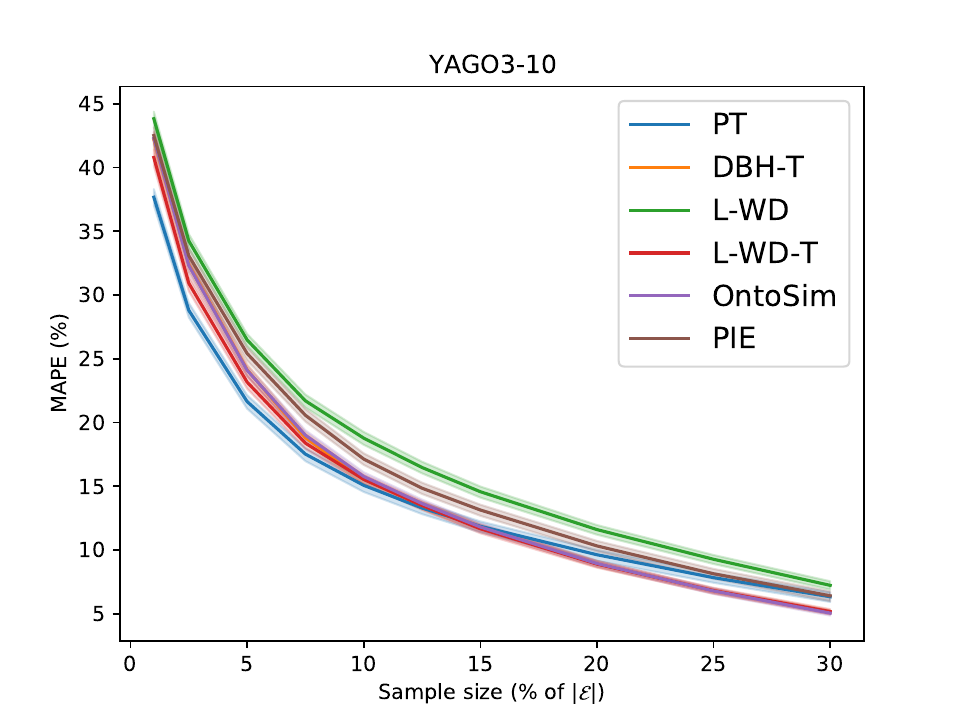}
    
   \label{fig:estimated-mrrs-ogbl-wikikg2}
    \end{subfigure}%
    
\caption{MAPE (\%) on FB15k, CoDEx-M and YAGO3-10. } \label{fig:2}
\end{figure*}

\paragraph{Relation Recommenders} Results displayed in Table \ref{table:candidate-recall-reltypes} show that L-WD performs similarly or better than PIE \cite{chao2022pie} at a fraction of the time and effort it takes to train it, as the training procedure for training the GNN is a more complex procedure. Moreover, PIE was trained on a GPU, while L-WD and L-WD-T only used sparse matrix multiplications on a CPU. As expected, the PT method cannot detect unseen candidates, hurting the CR in particular on large datasets. 

\paragraph{Choice of relational recommender and its impact on the estimation} In Figure \ref{fig:2}, we map the Mean Absolute Percentage Error (MAPE) against the maximum sample size and their 95 \% confidence intervals. For brevity, we display three datasets here (FB15k, CoDEx-M and YAGO3-10); the remaining can be seen in the Appendix, Section \ref{section:appendix-mape}. Overall, we observe that most of the time, all relational recommenders behave similarly with similar MAPE, although the MAPE estimates behave differently across datasets. PT is, expectedly, the method with the most amount of cases that does not converge towards a 0 \% MAPE. Otherwise, most methods typically converge quickly towards the true value at similar rates. Interestingly, the error of the estimates does not necessarily correlate with the performance of the methods on CR and RR; we find there to be a tendency of recommenders being "good enough" for most situations; as long as they catch enough candidates that are harder negatives, the estimation becomes similar in accuracy to other recommenders. We do not see a better estimation of the true performance metrics from using a sophisticated GCN model as PIE; but rather a similar performance to the simpler heuristics proposed. Which recommender performing the best tends to vary with each dataset; if types are available and of high quality, these can be used to produce L-WD-T, DBH-T or OntoSim, serving as strong heuristics; however, if types are not available, it can be important to have a method to resort to like L-WD.

\paragraph{Correlations and error rate} In Table \ref{table:correlations} we display the correlations and Table \ref{table:MAE-estimators} presents the MAEs in estimating the true metric. For brevity, we show the correlation between the true, filtered MRR and the different strategies; for comparisons with Hits@1, Hits@3 and Hits@10 we refer to the Appendix, Section \ref{sec:additional-results}. Interestingly, sampling uniformly at random does sometimes display a higher correlation, indicating that even sampling uniformly at random can be helpful if using the evaluation estimates for determining when to stop training. However, it is consistently outperformed in terms of MAE when estimating the actual metric (see Table \ref{table:MAE-estimators}).  %

$\mathcal{KP}$ demonstrates a certain instability in Table \ref{table:correlations}; at times, the correlation is negative across all sampling strategies (e.g., RESCAL on FB15k). This indicates that the method cannot always be used, and possibly risk that one chooses a worse performing KGC model if not used correctly. It is also important to note that none of the sampling strategies are entirely foolproof; training TransE on CoDEx demonstrates a situation where no estimation has a higher correlation than 0.786, and the MAE is also higher (at best, $0.049$ for Static sampling). However, $\mathcal{KP}$ does bring a better speed-up compared to estimating the true rank on the smaller datasets (see Table \ref{table:speed-up}). For brevity, the full results can be seen in the Appendix, Section \ref{section:speed-up}. Here, the relatively quick full evaluation on the smaller datasets leaves little room for improvement due to the small size of the datasets. Instead, the impact is much greater on ogbl-wikikg2, where we observe a speed-up of more than two orders of magnitude.

In the vast majority of our experiments, static sampling generally outperforms the other methods. However, on ogbl-wikikg2, we find that the most accurate is, by far, the probabilistic sampling (see Figure \ref{fig:estimated-mrrs-ogbl-wikikg2}) that coincides almost exactly with the true MRR while being on average 116 times faster than a full ranking estimate.

\paragraph{Kendall-Tau correlations} Table \ref{table:kendall-tau-correlations} displays the Kendall-Tau rank correlations, indicating how well each estimation preserves the true order of model performance when comparing the models. As can be seen, the Static sampling largely outperforms both Random and Probabilistic sampling, although Probabilistic does perform well. We conjecture that the lower performance of Random sampling here is largely explained due to the higher variance in the performance estimations, therefore increasing the uncertainty and changing positions of models in the performance order; in particular once the models have converged towards the true result. 

\paragraph{Large-scale evaluation} In Figure \ref{fig:1b-b} we display the sample size used versus the MRR estimation. Both our Probabilistic and Static sampling based on Relation Recommendation probabilities are significantly more accurate than random sampling and run faster than sampling uniformly at random, as shown in Figure \ref{fig:1a-b}. This improvement in terms of speed is partly caused by the reduced number of entities that occur with L-WD described in Section \ref{section:motivation}. Only by sampling approximately $2 \%$ of entities in our probabilistic setting, we achieve an accurate estimation of the true filtered MRR, beating the time needed by $\mathcal{KP}$ which is also very quick in its evaluation. We also see in Figure \ref{fig:1a-b} that the Static sampling grows slower in terms of time needed and increases non-linearly as expected. This demonstrates that an accurate selection of candidates does indeed greatly diminish the time needed, in particular when the size of the evaluation set is large. We see the same patterns on Hits@1, Hits@3 and Hits@10 (see Appendix, Section \ref{sec:hits-estimates}).

\section{Conclusion}

In this paper, we have shown that efficient methods for computing head- and tail set scores can guide the selection of counterexamples (negative samples) for evaluating a Knowledge Graph Completion model's capacity to predict candidate links to be added to Knowledge Graph, alleviating its heavy computational cost and simultaneously estimating the true performance of a model with a high level of reliability. The improvement is particularly noticeable in a large scale setting, implying that it is possible to estimate the true performance of a ranking model in significantly less time on large scale datasets which in turn allows quicker iterations on large scale Knowledge Graphs. For future work, we will investigate relation recommenders as a negative sample probabilities during training; while binary definitions of these have already been used by Krompass et al. \cite{krompass2015type} and Balkir et al. \cite{balkir-etal-2019-using}, a probabilistic recommendation of negative samples from a relation recommender remains. One can also investigate the use of easy negatives from scores being 0 in L-WD. While these are easy examples, it is an interesting direction as one can move to an almost guaranteed closed-world assumption. We will consider testing using these negative sampling domains in a setting to, for example, build a triplet classifier. 

Our framework is also extendable to other settings settings where the task is to rank entities, such as inductive KG completion \cite{teru2020inductive, lee2023ingram, Galkin2022} and logical query answering \cite{Ren*2020Query2box:, ren2020beta}, but is left for future work. Our sampling methods can also complement other metrics, such as ROC AUC and AUC-PR that have been used previously in KGC to better reflect a method's capability of predicting triples among harder examples, and we plan to investigate this. 

Ultimately, our framework enables an accurate and fast benchmarking of models, allowing for metrics to reflect the true performance of them. We hope that our work enables both faster prototyping and a higher reliability in the estimations of KGC model performance. 

\section{Acknowledgments}

This research was partially supported by the Wallenberg Autonomous Systems Program (WASP). The computations were enabled by the Berzelius resource provided by the Knut and Alice Wallenberg Foundation at the National Supercomputer Centre. We also thank Henrik Henriksson at NSC for his assistance with support on computational assistance, which was made possible through application support provided by the National Supercomputer Centre at Linköping University.

\bibliographystyle{ACM-Reference-Format}
\bibliography{cornelletal}
\appendix

\pagebreak
\newpage
\section{Proof of Theorem 1} \label{proof:theorem2}

In this section, we prove Theorem \ref{theorem:guaranteed}, i.e., that $\mathbb{E}[Y] \geq 0$.

Denote the set of entities that can be a tail of $r$ as the range of entities, i.e., $\mathcal{RS}_r = \{e_1, ..., e_{n_r}\}$. Given a typed ontology with given domains and ranges with types that are not noisy or ill-defined, we know that the $\mathcal{E}_{\mathcal{KG}_{(h,r)}} \subseteq \mathcal{RS}_r$. Now, given that we sample uniformly at random from this set, the distribution changes to $X_{\mathcal{E}_{(h,r)}} \sim \text{H}(|\mathcal{E}_{(h,r)}|,|\mathcal{RS}_r| - |\mathcal{E}_{(h,r)}|, n_s)$, with the restriction of $n_{s,r} = \min(n_s,|\mathcal{RS}_r|)$. %

For a query (h, r, ?), how much closer we land to the true rank can be estimated. As Y is distributed $Y \sim X_{\mathcal{R}_{r}} - X_{u}$, the expected number of ranks gained towards the true rank is the difference between $\mathbb{E}[X_{\mathcal{R}_{r}}]$ and $\mathbb{E}[X_{u}]$. 

\begin{align} \label{eq:gain}
    \mathbb{E}[Y] = \mathbb{E}[X_{\mathcal{R}_{r}} - X_{u}] = \\ \mathbb{E}[X_{\mathcal{R}_{r}}] - \mathbb{E}[X_{u}] = \\
    \frac{|\mathcal{E}_{(h,r)}|n_{s,r}}{|\mathcal{R}_r|} - \frac{|\mathcal{E}_{(h,r)}|n_s}{|\mathcal{E}|} = \\
    |\mathcal{E}_{(h,r)}|(\frac{\min(n_s, |\mathcal{RS}_r|)}{|\mathcal{R}_r|} - \frac{n_s}{|\mathcal{E}|})
\end{align}

Now, we have two cases; one where $\min(n_s, |\mathcal{RS}_r|) = n_s$ and $\min(n_s, |\mathcal{RS}_r|) = |\mathcal{RS}_r|$. In the first case, we get

\begin{align}
    |\mathcal{E}_{(h,r)}|\Big(\frac{n_s}{|\mathcal{RS}_r|} - \frac{n_s}{|\mathcal{E}|}) = \\
    |\mathcal{E}_{(h,r)}|n_s\Big(\frac{1}{|\mathcal{RS}_r|} - \frac{1}{|\mathcal{E}|}) = \\
    |\mathcal{E}_{(h,r)}|n_s\Big(\frac{|\mathcal{E}|-|\mathcal{RS}_r|}{|\mathcal{RS}_r||\mathcal{E}|}\Big)= \\
    \frac{|\mathcal{E}_{(h,r)}|n_s}{|\mathcal{RS}_r||\mathcal{E}|}\Big(|\mathcal{E}|-|\mathcal{RS}_r|\Big)
\end{align}

In the second case, we get

\begin{align}
    |\mathcal{E}_{(h,r)}|(\frac{|\mathcal{RS}_r|}{|\mathcal{RS}_r|} - \frac{n_s}{|\mathcal{E}|}) = \\
    |\mathcal{E}_{(h,r)}|(1 - \frac{n_s}{|\mathcal{E}|}) = \\
    \frac{|\mathcal{E}_{(h,r)}|}{|\mathcal{E}|}(|\mathcal{E}| - n_s)
\end{align}

Therefore, our given 

$$Y = \begin{cases}
        \frac{|\mathcal{E}_{(h,r)}|n_s}{|\mathcal{RS}_r||\mathcal{E}|}\Big(|\mathcal{E}|-|\mathcal{RS}_r|\Big) & \text{ if } n_s < \mathcal{RS}_r\\
        \frac{|\mathcal{E}_{(h,r)}|}{|\mathcal{E}|}(|\mathcal{E}| - n_s) & \text{otherwise}
    \end{cases}$$

The same can be proven for head queries, i.e., predicting heads or tails by considering $(?, r, t)$ as inverse $(h, r, ?)$ queries.

\begin{equation} \label{eq:firstcase}
    \frac{|\mathcal{E}_{(h,r)}|n_s}{|\mathcal{RS}_r||\mathcal{E}|}\Big(|\mathcal{E}|-|\mathcal{RS}_r|\Big) 
\end{equation}

Now, as Y is the expected gain of accuracy in estimating the rank, a negative value of Y would imply that we would be less accurate. Hence, to prove that sampling from the domains \& ranges give a more accurate rank estimation, we need to show that $Y \geq 0$ in all cases. Let's consider the first case, as in Equation \ref{eq:firstcase}. Here, we know that all values are natural numbers; the only determining factor making this possibly negative is $(|\mathcal{E}|-|\mathcal{RS}_r|)$. As we know that $\mathcal{RS}_r \subseteq \mathcal{E}$, this implies that $|\mathcal{RS}_r| \leq |\mathcal{E}|$, henceforth $|\mathcal{E}| - |\mathcal{RS}_r| \geq 0$, giving us the guarantee that $Y \geq 0$ when $n_s < \mathcal{RS}_r$. 

\begin{equation} \label{eq:secondcase}
    \frac{|\mathcal{E}_{(h,r)}|}{|\mathcal{E}|}(|\mathcal{E}| - n_s)
\end{equation}

Now, consider the second case as in Equation \ref{eq:secondcase}. As we know, $n_s \leq |\mathcal{E}|$ as $n_s$ is the number of items we sample from $\mathcal{E}$ without replacement. The same case is argued here; as $n_s \leq |\mathcal{E}|$, we get $|\mathcal{E}| - n_s \geq 0$, making $Y \geq 0$ when $n_s \geq \mathcal{RS}_r$. We have then proved that $Y \geq 0$, meaning that the gain in rank accuracy for an arbitrary query is equal or greater than 0 with our sampling scheme. This concludes the proof.

\section{Obtaining entity types} \label{sec:obtaining-entity-types}

For FB15k-237, we use the types from Wu et al.~\cite{wu2022learning}, and for YAGO3-10 we use the types from the YAGO website \footnote{\url{https://www.mpi-inf.mpg.de/departments/databases-and-information-systems/research/yago-naga/yago/downloads}}. For FB15k, we use the types used in numerours works on entity typing experiments, i.e., FB15kET \cite{moon2017learning}. For ogbl-wikikg2 and the CoDEx datasets we use the property \texttt{InstanceOf} (P31) to categorize the entities into types. While this is available for ogbl-wikikg2, we download these for the CoDEx datasets and release these along with our code. %

\section{Additional results} \label{sec:additional-results}

\begin{table*}[ht]
\caption{All false easy negatives generated by L-WD on three datasets. Note that there were no false hard negatives on YAGO3-10. }
\label{table:falsehardnegs}
\resizebox{\linewidth}{!}{%
\begin{tabular}{@{}clll@{}}
\toprule
\multicolumn{1}{l}{\textbf{Dataset}} & \textbf{Head}                                          & \textbf{Relation}                                                               & \textbf{Tail}                              \\ \midrule
\multirow{4}{*}{FB15k237}            & Azerbaijan                                             & /location/statistical\_region/religions./location/religion\_percentage/religion & Armenian Apostolic Church (/m/01hd99)      \\
                                     & Bhangra                                                & /common/topic/webpage./common/webpage/category                                  & /m/08mbj5d                                 \\
                                     & August                                                 & /people/person/gender                                                           & Male (/m/05zppz)                           \\
                                     & Stephen Harper                                         & /people/person/profession                                                       & /m/060m4                                   \\ \midrule
\multirow{35}{*}{ogbl-wikikg2}         & South American rugby Championships (Q2265547)          & sport (P641)                                                                    & rugby union (Q5849)                        \\
                                     & Deep Space Network (Q835696)                           & has use (P366)                                                                  & communications system (Q577764)            \\
                                     & Deep Space Network (Q835696)                           & physically interacts with (P129)                                                & spacecraft (Q40218)                        \\
                                     & Informalism (Q1054254)                                 & partially coincident with (P1382)                                               & abstract expressionism (Q177725)           \\
                                     & metronome (Q156424)                                    & has use (P366)                                                                  & tempo (Q189214)                            \\
                                     & Unknown (Q10638581)                                    & office contested (P541)                                                         & President of Finland (Q29558)              \\
                                     & Portrait of Alice Barnham (nee Bradbridge) (Q16687281) & P1773 (P1773)                                                                   & Hans Eworth (Q2468731)                     \\
                                     & gleba (Q2034230)                                       & part of (P361)                                                                  & mushroom (Q83093)                          \\
                                     & Apollo Lunar Module (Q208382)                          & part of (P361)                                                                  & Apollo space program (Q46611)              \\
                                     & Marc Étienne de Beauvau (Q3288546)                     & award received (P166)                                                           & Prince of the Holy Roman Empire (Q1411354) \\
                                     & interactive art (Q2394336)                             & field of work (P101)                                                            & installation art (Q212431)                 \\
                                     & Korean American (Q276879)                              & instance of (P31)                                                               & social group (Q874405)                     \\
                                     & Korean American (Q276879)                              & country of origin (P495)                                                        & Korea (Q18097)                             \\
                                     & Korean American (Q276879)                              & work location (P937)                                                            & United States of America (Q30)             \\
                                     & allegretto (Q2991873)                                  & instance of (P31)                                                               & tempo (Q189214)                            \\
                                     & Wisconsin Badgers (Q3098117)                           & represents (P1268)                                                              & University of Wisconsin–Madison (Q838330)  \\
                                     & politics in Afghanistan (Q1154652)                     & facet of (P1269)                                                                & Afghanistan (Q889)                         \\
                                     & Nintendo development teams (Q7039175)                  & is a list of (P360)                                                             & division (Q334453)                         \\
                                     & Nintendo development teams (Q7039175)                  & is a list of (P360)                                                             & company (Q783794)                          \\
                                     & heterochromatin (Q837783)                              & has part(s) (P527)                                                              & carbon (Q623)                              \\
                                     & Chromatin (Q180951)                                    & instance of (P31)                                                               & Chemical compound (Q11173)                 \\
                                     & Chromatin (Q180951)                                    & has part(s) (P527)                                                              & carbon (Q623)                              \\
                                     & Chromatin (Q180951)                                    & has part(s) (P527)                                                              & oxygen (Q629)                              \\
                                     & Italian Football Federation (Q201897)                  & head coach (P286)                                                               & Antonio Conte (Q26580)                     \\
                                     & 2014 Six Nations Championship (Q11176370)              & followed by (P156)                                                              & 2015 Six Nations Championship (Q11227920)  \\
                                     & Eric Caulier (Q3056233)                                & referee (P1652)                                                                 & Chinese martial arts (Q3705105)            \\
                                     & Eric Caulier (Q3056233)                                & dan/kyu rank (P468)                                                             & 6 (Q23488)                                 \\
                                     & ActRaiser (Q343033)                                    & review score by (P447)                                                          & IGN (Q207708)                              \\
                                     & Tony Thielemans (Q3531910)                             & dan/kyu rank (P468)                                                             & 6 (Q23488)                                 \\
                                     & Hironori Ōtsuka (Q386415)                              & dan/kyu rank (P468)                                                             & 10 (Q23806)                                \\
                                     & United States Constitution (Q11698)                    & approved by (P790)                                                              & Continental Congress (Q26718)              \\
                                     & Tatsuo Suzuki (Q529362)                                & dan/kyu rank (P468)                                                             & 8 (Q23355)                                 \\
                                     & Ber Borochov (Q720022)                                 & founded by (P112)                                                               & Labor Zionism (Q607470)                    \\
                                     & University of Wisconsin–Madison (Q838330)              & represented by (P1875)                                                          & Wisconsin Badgers (Q3098117)               \\
                                     & Nakayama Hakudō (Q1368811)                             & dan/kyu rank (P468)                                                             & 10 (Q23806)                                \\ \bottomrule
\end{tabular}%
}

\end{table*}

\subsection{Speed-up} \label{section:speed-up}

Table \ref{table:all-datasets-models-speedup} displays the fully detailed speed-ups for all datasets, models and sampling techniques, including the average runtime with their standard deviations across the 100 epochs. 

\begin{table*}[]
\caption{Average speed-up (with standard deviations) on smaller datasets with different models (higher is better). }
\label{table:all-datasets-models-speedup}
\begin{tabular}{@{}crrrr|rrr|c@{}}
\toprule
\multicolumn{1}{l}{}                & \multicolumn{1}{l}{}               & \multicolumn{3}{c|}{$\mathcal{KP}$}                                                                  & \multicolumn{3}{c|}{\textbf{Sampling}}                                                            & \multicolumn{1}{l}{}                                        \\ \midrule
\textbf{Dataset}                    & \multicolumn{1}{c}{\textbf{Model}} & \multicolumn{1}{c}{\textbf{R}} & \multicolumn{1}{c}{\textbf{P}} & \multicolumn{1}{c|}{\textbf{S}} & \multicolumn{1}{c}{\textbf{R}} & \multicolumn{1}{c}{\textbf{P}} & \multicolumn{1}{c|}{\textbf{S}} & \multicolumn{1}{l}{\textbf{Full evaluation time (seconds)}} \\ \midrule
\multirow{6}{*}{\textbf{FB15k-237}} & \textbf{ComplEx}                   & $0.8 \pm 0.1$                  & $0.7 \pm 0.1$                  & $0.7 \pm 0.1$                   & $0.4 \pm 0.0$                  & $0.3 \pm 0.0$                  & $0.3 \pm 0.0$                   & $0.7 \pm 0.1$                                               \\
                                    & \textbf{ConvE}                     & $7.7 \pm 0.5$                  & $5.6 \pm 0.5$                  & $5.5 \pm 0.5$                   & $3.1 \pm 0.2$                  & $2.8 \pm 0.2$                  & $2.8 \pm 0.2$                   & $7.1 \pm 0.3$                                               \\
                                    & \textbf{DistMult}                  & $8.2 \pm 0.4$                  & $5.9 \pm 0.6$                  & $5.8 \pm 0.6$                   & $3.6 \pm 0.2$                  & $3.2 \pm 0.2$                  & $3.1 \pm 0.3$                   & $6.7 \pm 0.3$                                               \\
                                    & \textbf{RESCAL}                    & $7.9 \pm 0.2$                  & $5.9 \pm 0.6$                  & $5.8 \pm 0.6$                   & $3.6 \pm 0.1$                  & $3.2 \pm 0.1$                  & $3.2 \pm 0.2$                   & $6.7 \pm 0.2$                                               \\
                                    & \textbf{RotatE}                    & $12.4 \pm 1.0$                 & $9.1 \pm 1.1$                  & $8.6 \pm 1.1$                   & $4.8 \pm 0.3$                  & $4.4 \pm 0.3$                  & $4.2 \pm 0.4$                   & $10.0 \pm 0.6$                                              \\
                                    & \textbf{TransE}                    & $10.2 \pm 0.8$                 & $7.5 \pm 1.0$                  & $7.0 \pm 0.8$                   & $4.4 \pm 0.3$                  & $4.0 \pm 0.3$                  & $3.8 \pm 0.3$                   & $8.1 \pm 0.5$                                               \\ \midrule
\multirow{6}{*}{\textbf{FB15k}}     & \textbf{ComplEx}                   & $15.1 \pm 0.3$                 & $11.5 \pm 0.8$                 & $11.5 \pm 0.8$                  & $2.6 \pm 0.1$                  & $2.2 \pm 0.1$                  & $2.2 \pm 0.1$                   & $19.2 \pm 0.3$                                              \\
                                    & \textbf{ConvE}                     & $12.0 \pm 0.2$                 & $9.8 \pm 0.7$                  & $9.8 \pm 0.7$                   & $2.4 \pm 0.1$                  & $2.1 \pm 0.1$                  & $2.1 \pm 0.1$                   & $20.7 \pm 0.2$                                              \\
                                    & \textbf{DistMult}                  & $14.9 \pm 0.6$                 & $12.3 \pm 0.4$                 & $12.3 \pm 0.3$                  & $2.7 \pm 0.1$                  & $2.3 \pm 0.1$                  & $2.3 \pm 0.1$                   & $19.1 \pm 0.3$                                              \\
                                    & \textbf{RESCAL}                    & $12.9 \pm 0.2$                 & $11.0 \pm 0.2$                 & $10.9 \pm 0.2$                  & $2.7 \pm 0.2$                  & $2.3 \pm 0.1$                  & $2.3 \pm 0.1$                   & $19.2 \pm 0.2$                                              \\
                                    & \textbf{RotatE}                    & $22.3 \pm 1.0$                 & $18.2 \pm 0.7$                 & $18.2 \pm 0.7$                  & $3.5 \pm 0.2$                  & $3.1 \pm 0.1$                  & $3.0 \pm 0.1$                   & $28.1 \pm 0.4$                                              \\
                                    & \textbf{TransE}                    & $18.6 \pm 0.4$                 & $15.3 \pm 0.3$                 & $15.2 \pm 0.4$                  & $3.3 \pm 0.1$                  & $2.8 \pm 0.1$                  & $2.8 \pm 0.1$                   & $23.5 \pm 0.3$                                              \\ \midrule
\multirow{4}{*}{\textbf{CoDEx-S}}   & \textbf{ComplEx}                   & $10.1 \pm 1.0$                 & $8.5 \pm 2.4$                  & $8.9 \pm 1.5$                   & $2.7 \pm 0.2$                  & $2.4 \pm 0.3$                  & $2.4 \pm 0.3$                   & $0.7 \pm 0.1$                                               \\
                                    & \textbf{ConvE}                     & $8.4 \pm 0.6$                  & $7.4 \pm 2.0$                  & $8.2 \pm 1.4$                   & $2.4 \pm 0.2$                  & $2.2 \pm 0.3$                  & $2.2 \pm 0.2$                   & $0.7 \pm 0.1$                                               \\
                                    & \textbf{RESCAL}                    & $5.9 \pm 0.6$                  & $5.2 \pm 1.1$                  & $5.5 \pm 0.6$                   & $2.8 \pm 0.2$                  & $2.5 \pm 0.2$                  & $2.5 \pm 0.3$                   & $0.7 \pm 0.1$                                               \\
                                    & \textbf{TransE}                    & $10.3 \pm 1.1$                 & $8.4 \pm 2.6$                  & $9.1 \pm 1.5$                   & $3.0 \pm 0.2$                  & $2.6 \pm 0.3$                  & $2.5 \pm 0.3$                   & $0.7 \pm 0.1$                                               \\ \midrule
\multirow{2}{*}{\textbf{CoDEx-M}}   & \textbf{ComplEx}                   & $8.7 \pm 0.8$                  & $4.6 \pm 0.4$                  & $4.6 \pm 0.5$                   & $6.1 \pm 0.4$                  & $5.6 \pm 0.5$                  & $5.3 \pm 0.4$                   & $4.1 \pm 0.3$                                               \\
                                    & \textbf{ConvE}                     & $6.8 \pm 0.7$                  & $4.2 \pm 0.4$                  & $4.2 \pm 0.5$                   & $5.7 \pm 0.5$                  & $5.4 \pm 0.5$                  & $5.0 \pm 0.4$                   & $4.3 \pm 0.4$                                               \\ \midrule
\multirow{5}{*}{\textbf{CoDEx-L}}   & \textbf{TuckER}                    & $5.5 \pm 0.1$                  & $2.8 \pm 0.1$                  & $2.8 \pm 0.1$                   & $7.2 \pm 0.1$                  & $6.9 \pm 0.2$                  & $5.4 \pm 0.1$                   & $13.6 \pm 0.3$                                              \\
                                    & \textbf{ComplEx}                   & $6.7 \pm 0.1$                  & $3.1 \pm 0.1$                  & $3.1 \pm 0.1$                   & $7.4 \pm 0.1$                  & $7.0 \pm 0.1$                  & $5.5 \pm 0.1$                   & $13.6 \pm 0.2$                                              \\
                                    & \textbf{ConvE}                     & $6.8 \pm 0.4$                  & $3.2 \pm 0.1$                  & $3.2 \pm 0.2$                   & $6.9 \pm 0.3$                  & $6.5 \pm 0.3$                  & $5.3 \pm 0.2$                   & $14.4 \pm 0.5$                                              \\
                                    & \textbf{RESCAL}                    & $6.6 \pm 0.2$                  & $3.0 \pm 0.1$                  & $3.0 \pm 0.1$                   & $7.8 \pm 0.4$                  & $7.3 \pm 0.2$                  & $5.8 \pm 0.1$                   & $13.3 \pm 0.2$                                              \\
                                    & \textbf{TransE}                    & $11.8 \pm 0.4$                 & $5.4 \pm 0.2$                  & $5.4 \pm 0.2$                   & $8.7 \pm 0.5$                  & $9.6 \pm 0.6$                  & $4.5 \pm 0.2$                   & $23.2 \pm 0.7$                                              \\ \midrule
\textbf{YAGO3-10}                   & \textbf{ComplEx}                   & $4.4 \pm 0.3$                  & $0.9 \pm 0.2$                  & $0.9 \pm 0.2$                   & $2.3 \pm 0.1$                  & $1.9 \pm 0.2$                  & $1.6 \pm 0.2$                   & $1.2 \pm 0.1$                                               \\ \bottomrule
\end{tabular}
\end{table*}

\begin{table*}[]
\caption{Correlation with the filtered Hits@3. Best estimators in bold. }
\begin{tabular}{cc|rrr|rrr}
\hline
\multicolumn{1}{l}{\textbf{}}       & \textbf{}         & \multicolumn{3}{c|}{$\mathcal{KP}$}                                                                           & \multicolumn{3}{c}{\textbf{Sampling}}                                                            \\ \hline
\textbf{Dataset}                    & \textbf{Model}    & \multicolumn{1}{c}{\textbf{R}} & \multicolumn{1}{c}{\textbf{P}} & \multicolumn{1}{c|}{\textbf{S}} & \multicolumn{1}{c}{\textbf{R}} & \multicolumn{1}{c}{\textbf{P}} & \multicolumn{1}{c}{\textbf{S}} \\ \hline
\multirow{6}{*}{\textbf{FB15k-237}} & \textbf{TransE}   & -0.232                         & -0.479                         & -0.446                          & \textbf{0.997}                 & 0.989                          & 0.980                          \\
                                    & \textbf{RotatE}   & 0.514                          & 0.069                          & -0.099                          & \textbf{0.997}                 & 0.995                          & 0.986                          \\
                                    & \textbf{RESCAL}   & 0.050                          & 0.792                          & 0.741                           & \textbf{0.995}                 & \textbf{0.995}                 & 0.994                          \\
                                    & \textbf{DistMult} & 0.909                          & 0.843                          & 0.855                           & \textbf{0.997}                 & 0.993                          & 0.994                          \\
                                    & \textbf{ConvE}    & 0.515                          & 0.802                          & 0.754                           & 0.995                          & \textbf{0.999}                 & 0.998                          \\
                                    & \textbf{ComplEx}  & 0.896                          & 0.839                          & 0.846                           & 0.995                          & \textbf{0.996}                 & 0.992                          \\ \hline
\multirow{6}{*}{\textbf{FB15k}}     & \textbf{TransE}   & 0.013                          & 0.107                          & 0.025                           & 0.989                          & \textbf{1.000}                 & \textbf{1.000}                 \\
                                    & \textbf{RotatE}   & -0.071                         & 0.290                          & 0.150                           & 0.964                          & \textbf{1.000}                 & 0.999                          \\
                                    & \textbf{RESCAL}   & -0.134                         & -0.382                         & -0.340                          & 0.991                          & \textbf{1.000}                 & \textbf{1.000}                 \\
                                    & \textbf{DistMult} & 0.110                          & 0.577                          & 0.532                           & 0.969                          & \textbf{1.000}                 & \textbf{1.000}                 \\
                                    & \textbf{ConvE}    & 0.897                          & 0.943                          & 0.932                           & 0.984                          & \textbf{1.000}                 & 0.999                          \\
                                    & \textbf{ComplEx}  & 0.091                          & 0.185                          & 0.194                           & 0.977                          & 0.999                          & \textbf{1.000}                 \\ \hline
\multirow{4}{*}{\textbf{CoDEx-S}}   & \textbf{TransE}   & 0.146                          & 0.610                          & 0.555                           & 0.975                          & 0.993                          & \textbf{0.999}                 \\
                                    & \textbf{RESCAL}   & -0.214                         & 0.123                          & 0.028                           & 0.829                          & 0.985                          & \textbf{0.999}                 \\
                                    & \textbf{ConvE}    & -0.191                         & 0.419                          & 0.304                           & 0.937                          & 0.995                          & \textbf{1.000}                 \\
                                    & \textbf{ComplEx}  & 0.700                          & 0.888                          & 0.908                           & 0.975                          & 0.995                          & \textbf{0.997}                 \\ \hline
\multirow{2}{*}{\textbf{CoDEx-M}}   & \textbf{ConvE}    & 0.771                          & 0.760                          & 0.743                           & 0.980                          & 0.999                          & \textbf{1.000}                 \\
                                    & \textbf{ComplEx}  & 0.404                          & 0.689                          & 0.780                           & 0.977                          & 0.997                          & \textbf{1.000}                 \\ \hline
\multirow{5}{*}{\textbf{CoDEx-L}}   & \textbf{TransE}   & 0.005                          & 0.264                          & 0.356                           & 0.894                          & 0.942                          & \textbf{0.951}                 \\
                                    & \textbf{TuckER}   & -0.172                         & 0.262                          & 0.192                           & 0.933                          & 0.996                          & \textbf{1.000}                 \\
                                    & \textbf{RESCAL}   & 0.340                          & 0.254                          & 0.278                           & 0.952                          & 0.998                          & \textbf{1.000}                 \\
                                    & \textbf{ConvE}    & 0.872                          & 0.852                          & 0.852                           & 0.989                          & 0.998                          & \textbf{1.000}                 \\
                                    & \textbf{ComplEx}  & -0.134                         & 0.082                          & 0.168                           & 0.493                          & \textbf{0.989}                 & 0.987                          \\ \hline
\textbf{YAGO3-10}                   & \textbf{ComplEx}  & 0.513                          & 0.618                          & 0.657                           & 0.961                          & 0.938                          & \textbf{0.979}                 \\ \hline
\end{tabular}
\end{table*}

\begin{table*}[]
\caption{Correlation with the Filtered Validation on Hits@10. Bold indicates best estimator.}

\begin{tabular}{cc|rrr|rrr}
\toprule
\multicolumn{1}{l}{\textbf{}}       & \textbf{}         & \multicolumn{3}{c|}{$\mathcal{KP}$}                                                                           & \multicolumn{3}{c}{\textbf{Sampling}}                                                            \\
\midrule
\textbf{Dataset}                    & \textbf{Model}    & \multicolumn{1}{c}{\textbf{R}} & \multicolumn{1}{c}{\textbf{P}} & \multicolumn{1}{c|}{\textbf{S}} & \multicolumn{1}{c}{\textbf{R}} & \multicolumn{1}{c}{\textbf{P}} & \multicolumn{1}{c}{\textbf{S}} \\ \hline
\multirow{6}{*}{\textbf{FB15k-237}}& \textbf{TransE}               & -0.236                & -0.479                & -0.446                & \textbf{0.993}                 & 0.984                 & 0.971                 \\
                           & \textbf{RotatE}               & 0.516                 & 0.069                 & -0.095                & \textbf{0.998}                 & 0.994                 & 0.983                 \\
                           & \textbf{RESCAL}               & 0.089                 & 0.791                 & 0.743                 & 0.993                 &\textbf{ 0.995}                 & 0.991                 \\
                           & \textbf{DistMult}             & 0.905                 & 0.835                 & 0.849                 & \textbf{0.998}                 & 0.993                 & 0.992                 \\
                           & \textbf{ConvE}                & 0.520                 & 0.804                 & 0.753                 & 0.995                 & \textbf{0.999}                 & 0.998                 \\
                           & \textbf{ComplEx}              & 0.890                 & 0.833                 & 0.843                 & 0.995                 & \textbf{0.997}                 & 0.988                 \\
                           \midrule
\multirow{6}{*}{\textbf{FB15k}}    & \textbf{TransE}               & -0.004                & 0.120                 & 0.022                 & 0.972                 & 0.999                 & \textbf{1.000}                 \\
                           & \textbf{RotatE}               & -0.072                & 0.266                 & 0.144                 & 0.963                 & \textbf{0.999}                 & \textbf{0.999}                 \\
                           & \textbf{RESCAL}               & -0.112                & -0.366                & -0.323                & 0.985                 & 0.999                 & \textbf{1.000}                 \\
                           & \textbf{DistMult}             & 0.098                 & 0.552                 & 0.498                 & 0.964                 & \textbf{1.000}                 & \textbf{1.000}                 \\
                           & \textbf{ConvE}                & 0.911                 & 0.941                 & 0.934                 & 0.969                 & \textbf{0.999}                 & 0.998                 \\
                           & \textbf{ComplEx}              & 0.102                 & 0.198                 & 0.198                 & 0.968                 & 0.998                 & \textbf{1.000}                 \\
                           \midrule
\multirow{4}{*}{\textbf{CoDEx-S}}   & \textbf{TransE}               & 0.144                 & 0.612                 & 0.563                 & 0.986                 & 0.997                 & \textbf{1.000}                 \\
                           & \textbf{RESCAL}               & -0.138                & 0.143                 & 0.036                 & 0.822                 & 0.976                 & \textbf{0.998}                 \\
                           & \textbf{ConvE}                & -0.156                & 0.443                 & 0.316                 & 0.927                 & 0.995                 & \textbf{1.000}                 \\
                           & \textbf{ComplEx}              & 0.714                 & 0.898                 & 0.920                 & 0.976                 & 0.991                 & \textbf{0.993}                 \\
                           \midrule
\multirow{2}{*}{\textbf{CoDEx-M}}   & \textbf{ConvE}                & 0.757                 & 0.747                 & 0.728                 & 0.995                 & 0.999                 & \textbf{1.000}                 \\
                           & \textbf{ComplEx}              & 0.397                 & 0.684                 & 0.785                 & 0.995                 & \textbf{0.999}                 & \textbf{0.999}                 \\
                           \midrule
\multirow{5}{*}{\textbf{CoDEx-L}}   & \textbf{TransE}               & -0.008                & 0.373                 & 0.460                 & 0.967                 & 0.975                 & \textbf{0.980}                 \\
                           & \textbf{TuckER}    & -0.149                & 0.275                 & 0.214                 & 0.984                 & 0.996                 & \textbf{1.000}                 \\
                           & \textbf{RESCAL}               & 0.334                 & 0.271                 & 0.300                 & 0.983                 & 0.996                 & \textbf{1.000}                 \\
                           & \textbf{ConvE}                & 0.873                 & 0.854                 & 0.855                 & 0.999                 & 0.998                 & \textbf{1.000}                 \\
                           & \textbf{ComplEx}              & -0.016                & 0.189                 & 0.270                 & 0.817                 & \textbf{0.975}                 & 0.958                 \\
                           \midrule
\textbf{YAGO3-10}                  & \textbf{ComplEx}              & 0.486                 & 0.595                 & 0.645                 & 0.962                 & 0.960                 & \textbf{0.980}                \\
\bottomrule
\end{tabular}
\end{table*}

\subsection{False negatives produced by L-WD} \label{sec:false-negs-lWD}

Table \ref{table:falsehardnegs} displays all the false negatives of different datasets on which the investigation was conducted. As we can see, several triples are unintuitive and can be attributed to errors during the construction. 
\begin{table*}[]
\caption{Correlations with the Filtered Validation on Hits@1. Bold indicates best estimator.}

\begin{tabular}{cc|rrr|rrr}
\toprule
\multicolumn{1}{l}{\textbf{}}       & \textbf{}         & \multicolumn{3}{c|}{$\mathcal{KP}$}                                                                           & \multicolumn{3}{c}{\textbf{Sampling}}                                                            \\
\midrule
\textbf{Dataset}                    & \textbf{Model}    & \multicolumn{1}{c}{\textbf{R}} & \multicolumn{1}{c}{\textbf{P}} & \multicolumn{1}{c|}{\textbf{S}} & \multicolumn{1}{c}{\textbf{R}} & \multicolumn{1}{c}{\textbf{P}} & \multicolumn{1}{c}{\textbf{S}} \\ \hline
\multirow{6}{*}{\textbf{FB15k-237}}& \textbf{TransE}                    & -0.229                & -0.477                & -0.451                &  \textbf{0.992}                 & 0.982                 & 0.974                 \\
                           & \textbf{RotatE}                    & 0.512                 & 0.069                 & -0.094                & 0.991                 & \textbf{0.992}                 & 0.983                 \\
                           & \textbf{RESCAL}                    & 0.041                 & 0.793                 & 0.743                 & 0.993                 & 0.994                 & \textbf{0.996}                 \\
                           & \textbf{DistMult}                  & 0.910                 & 0.843                 & 0.856                 & 0.993                 & 0.994                 & \textbf{0.995}                 \\
                           & \textbf{ConvE}                     & 0.512                 & 0.803                 & 0.756                 & 0.981                 & \textbf{0.999}                 & \textbf{0.999}                 \\
                           & \textbf{ComplEx}                   & 0.903                 & 0.843                 & 0.851                 & 0.989                 & \textbf{0.997}                 & 0.994                 \\
                           \midrule
\multirow{6}{*}{\textbf{FB15k}}    & \textbf{TransE}                    & 0.046                 & 0.090                 & 0.030                 & 0.990                 & 0.999                 & \textbf{1.000}                 \\
                           & \textbf{RotatE}                    & -0.066                & 0.310                 & 0.157                 & 0.972                 & \textbf{1.000}                 & 0.999                 \\
                           & \textbf{RESCAL}                    & -0.150                & -0.400                & -0.355                & 0.992                 & \textbf{1.000}                 & \textbf{1.000}                 \\
                           & \textbf{DistMult}                  & 0.128                 & 0.595                 & 0.561                 & 0.972                 & \textbf{1.000}                 & \textbf{1.000}                 \\
                           & \textbf{ConvE}                     & 0.876                 & 0.932                 & 0.919                 & 0.987                 & \textbf{1.000}                 & \textbf{1.000}                 \\
                           & \textbf{ComplEx}                   & 0.076                 & 0.170                 & 0.192                 & 0.985                 & \textbf{1.000}                 & \textbf{1.000}                 \\
                           \midrule
\multirow{4}{*}{\textbf{CoDEx-S}}   & \textbf{TransE}                    & 0.122                 & 0.571                 & 0.538                 & 0.866                 & 0.947                 &  \textbf{0.998}                 \\
                           & \textbf{RESCAL}                    & -0.183                & 0.118                 & 0.038                 & 0.806                 & 0.968                 & \textbf{1.000}                 \\
                           & \textbf{ConvE}                     & -0.190                & 0.428                 & 0.255                 & 0.921                 & 0.988                 & \textbf{1.000}                 \\
                           & \textbf{ComplEx}                   & 0.683                 & 0.872                 & 0.892                 & 0.977                 & 0.996                 & \textbf{0.999}                 \\
                           \midrule
\multirow{2}{*}{\textbf{CoDEx-M}}   & \textbf{ConvE}                     & 0.782                 & 0.769                 & 0.753                 & 0.927                 & 0.999                 & \textbf{1.000}                 \\
                           & \textbf{ComplEx}                   & 0.404                 & 0.680                 & 0.778                 & 0.933                 & 0.995                 & \textbf{1.000}                 \\
                           \midrule
\multirow{5}{*}{\textbf{CoDEx-L}}   & \textbf{TransE}                    & 0.028                 & -0.111                & -0.026                & \textbf{0.640}                 & 0.516                 & 0.531                 \\
                           & \textbf{TuckER}         & -0.195                & 0.256                 & 0.174                 & 0.787                 & 0.993                 & \textbf{1.000}                 \\
                           & \textbf{RESCAL}                    & 0.352                 & 0.229                 & 0.258                 & 0.863                 & 0.998                 & \textbf{1.000}                 \\
                           & \textbf{ConvE}                     & 0.869                 & 0.850                 & 0.849                 & 0.956                 & 0.999                 & \textbf{1.000}                 \\
                           & \textbf{ComplEx}                   & -0.209                & -0.012                & 0.056                 & 0.267                 & 0.996                 & \textbf{0.997}                 \\
                           \midrule
\textbf{YAGO3-10}                   & \textbf{ComplEx}                   & 0.547                 & 0.647                 & 0.696                 & 0.979                 & 0.960                 & \textbf{0.986}                \\
\bottomrule
\end{tabular}
\end{table*}

\subsection{MAPE on remaining datasets} \label{section:appendix-mape}

\begin{figure*}[ht] 
\caption{MAPE (\%) on FB15k-237 (left) and CoDEx-M (right)} \label{fig:1}
  \begin{subfigure}{0.5\textwidth}
  \caption{MAPE on FB15k-237. } \label{fig:1a-a}
    \includegraphics[width=\linewidth]{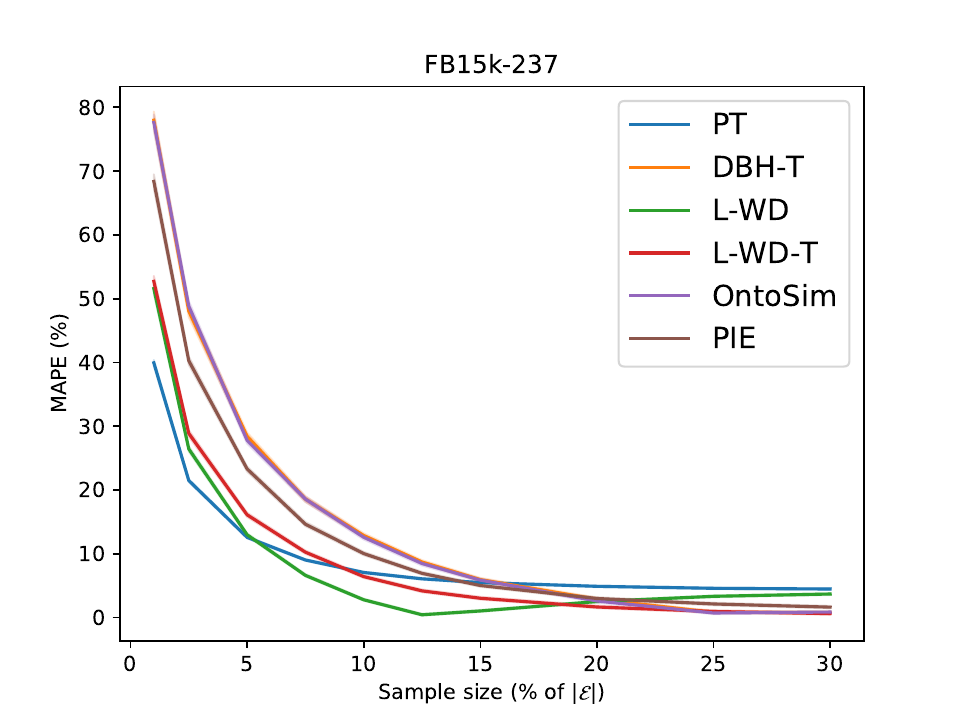}
    
  \end{subfigure}%
  \hspace*{\fill}   %
  \begin{subfigure}{0.5\textwidth}
  \caption{CoDEx-L. } \label{fig:1b-b}
    \includegraphics[width=\linewidth]{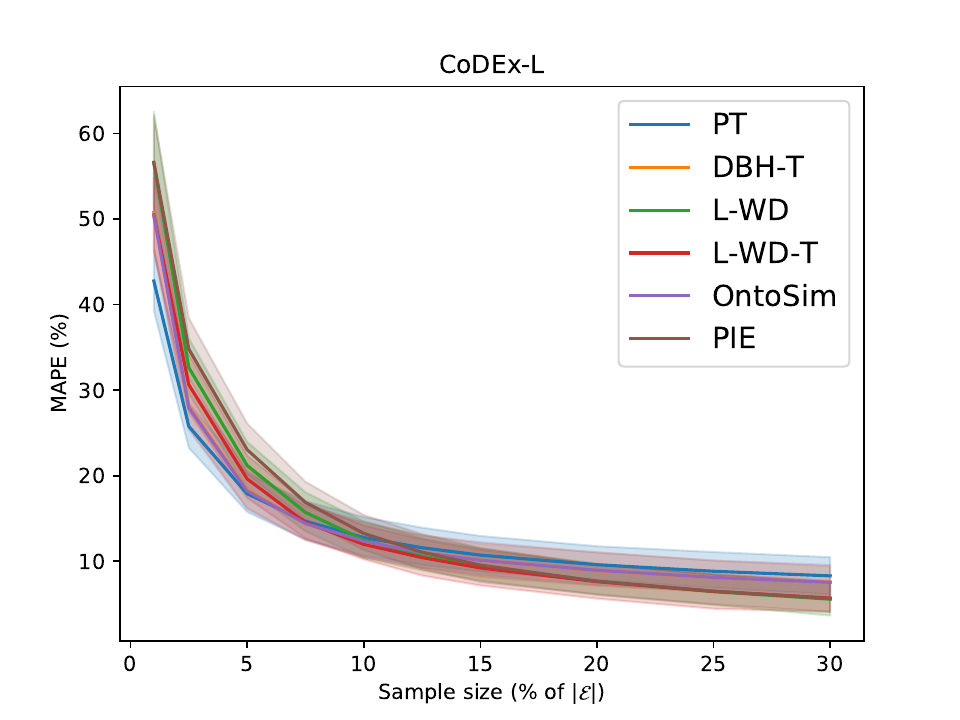}
    
  \end{subfigure}%
    \hspace*{\fill}   %

\begin{subfigure}{0.5\textwidth}
  \caption{ogbl-wikikg2. } \label{fig:1a-b}
    \includegraphics[width=\linewidth]{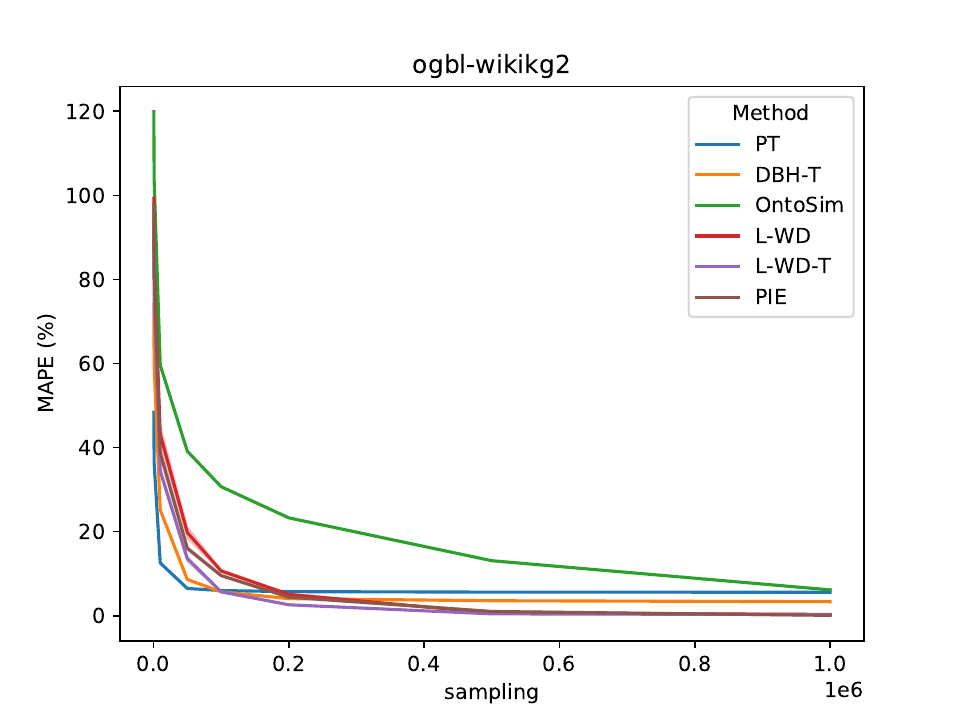}
    
  \end{subfigure}%
  \hspace*{\fill}   %
  \begin{subfigure}{0.5\textwidth}
  \caption{CoDEx-S. } \label{fig:1b-c}
    \includegraphics[width=\linewidth]{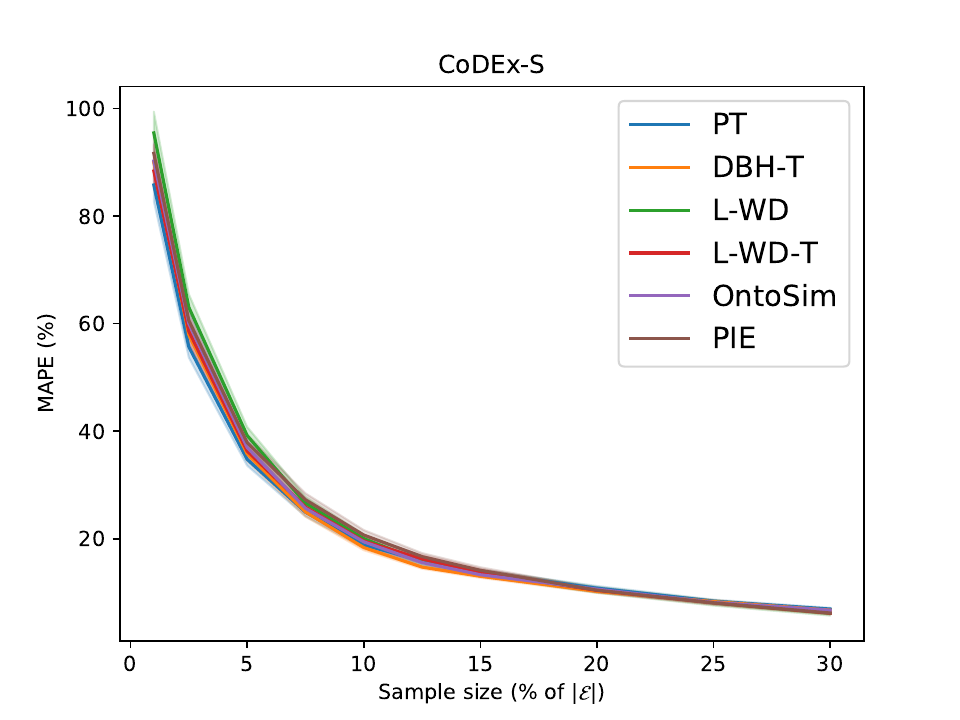}
    
  \end{subfigure}%
    \hspace*{\fill}   %

\end{figure*}

For MAPE on the remaining datasets, we refer to figures \ref{fig:1a-a}, \ref{fig:1b-b} and \ref{fig:1b-c}.

\subsection{Estimations of Hits@1, Hits@3 and Hits@10 on ogbl-wikikg2} \label{sec:hits-estimates}

\begin{figure*}[ht]
\caption{Hits@X metrics mapped against the sample size. } \label{fig:3}
\begin{subfigure}{0.33\textwidth}
  \caption{Hits@1 vs. sample size. } \label{fig:3ax}
    \centering
    \includegraphics[width=\linewidth]{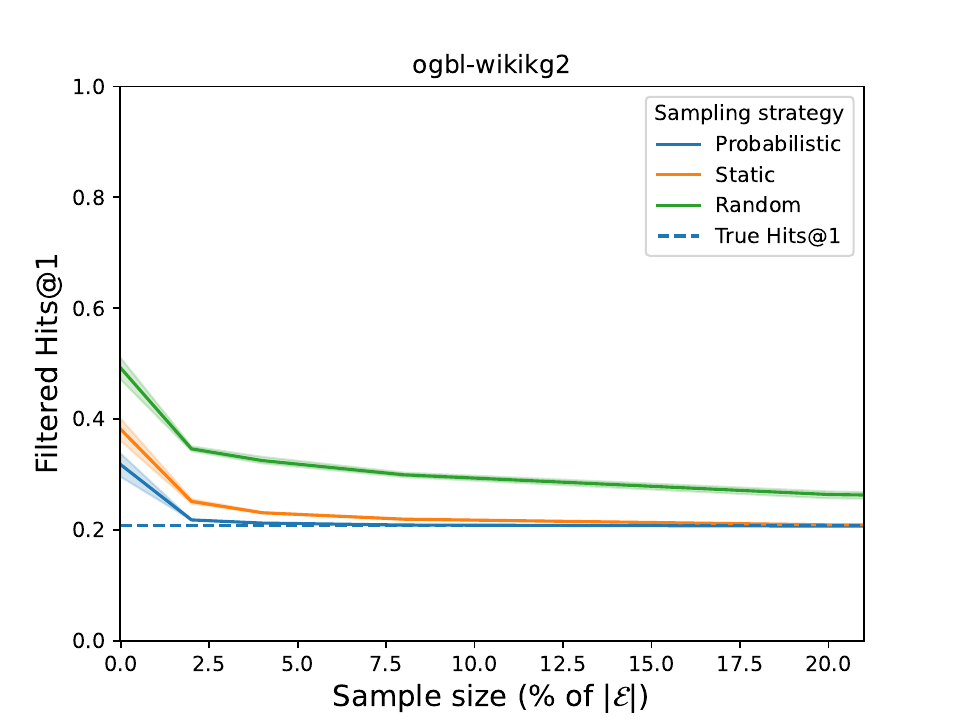}
    
  \end{subfigure}%
  \hspace*{\fill}   %
  \begin{subfigure}{0.33\textwidth}
    \caption{Hits@3 vs. sample size. } \label{fig:3bx}
    \centering
    \includegraphics[width=\linewidth]{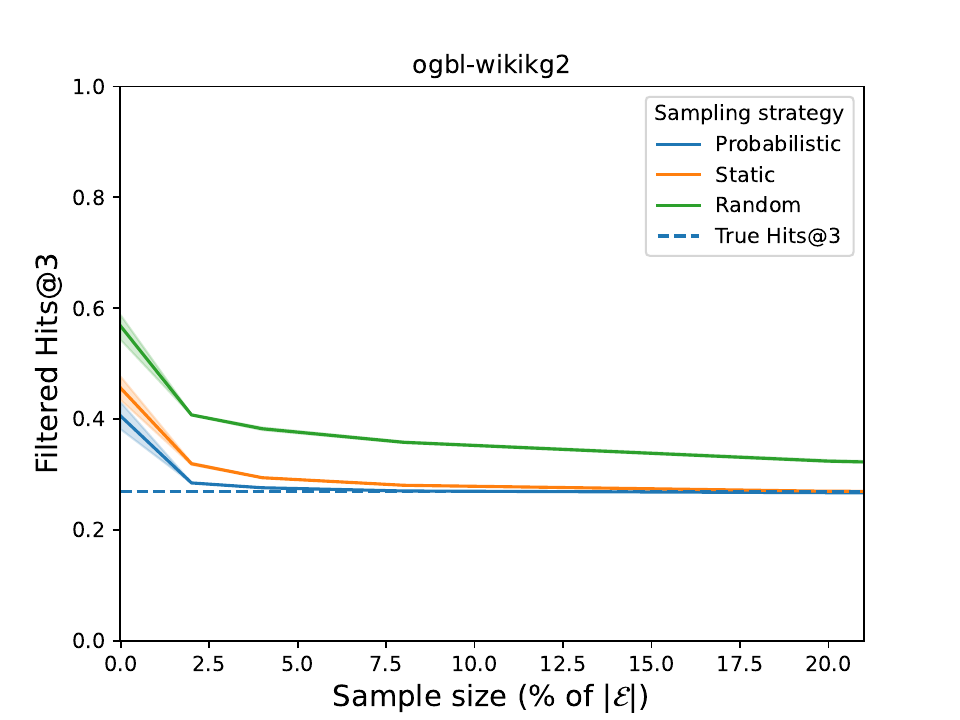}
    
  \end{subfigure}%
    \hspace*{\fill}   %
    \begin{subfigure}{0.33\textwidth}

   \caption{Hits@10 vs. sample size. } \label{fig:3cx}
    \centering
    \includegraphics[width=\linewidth]{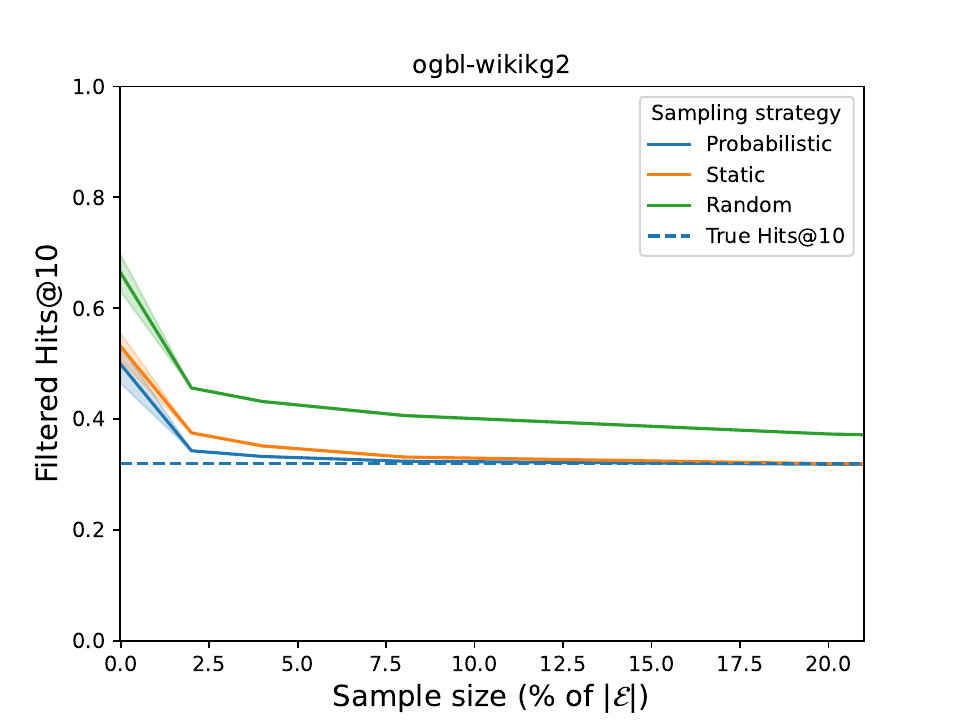}
    \end{subfigure}%

\end{figure*}

In Figure \ref{fig:3ax}, \ref{fig:3bx} and \ref{fig:3cx}, we display the estimates of the different methods of Hits@1, Hits@3 and Hits@10. We see the same pattern here as for the Filtered MRR.

\section{Training specifications of PIE} \label{sec:pie-training}

For PIE, we adopted their code\footnote{\url{https://github.com/ant-research/Parameter\_Inference\_Efficient\_PIE}} trained using a hidden dimensionality of 100, with Adam \cite{kingma2014adam} as optimizer and a learning rate of 0.0001. We use a margin of 3 and a batch size of 10. Several of these were the default settings; we increased the dimensionality from 64 as the original settings were intended for WikiKG90Mv2\footnote{\url{https://ogb.stanford.edu/docs/lsc/leaderboards/\#wikikg90mv2}}. We set early stopping for 20 epochs on FB15k237 and YAGO310, and 100 for ogbl-wikikg2. 

\section{Hardware specifications}

All experiments except for the evaluation of the pretrained ComplEx embeddings from Chen et al. \cite{chen2021relation} are run on a RTX2390 16 GB GPU on a computer with 188 GB RAM and 24 cores (48 virtual). The evaluation of the embeddings on ogbl-wikikg2 are run on an A100 40 GB GPU with a 256 GB RAM. Both of these devices were Linux OS. 

\begin{table*}[]
\caption{MAEs of estimating the true rank of Hits@X metrics. }
\begin{tabular}{@{}ll|lll|lll|lll@{}}
\toprule
                                                      &                                     & \multicolumn{3}{c|}{\textbf{Hits@1}}                                                              & \multicolumn{3}{c|}{\textbf{Hits@3}}                                                              & \multicolumn{3}{c}{\textbf{Hits@10}}                                                              \\ \midrule
\multicolumn{1}{c}{\textbf{Dataset}} & \multicolumn{1}{c|}{\textbf{Model}} & \multicolumn{1}{c}{\textbf{P}} & \multicolumn{1}{c}{\textbf{R}} & \multicolumn{1}{c|}{\textbf{S}} & \multicolumn{1}{c}{\textbf{P}} & \multicolumn{1}{c}{\textbf{R}} & \multicolumn{1}{c|}{\textbf{S}} & \multicolumn{1}{c}{\textbf{P}} & \multicolumn{1}{c}{\textbf{R}} & \multicolumn{1}{c}{\textbf{S}}  \\ \midrule
\multirow{6}{*}{\textbf{FB15k-237}}                   & \textbf{TransE}    & 0,013                          & 0,197                          & \textbf{0,008} & 0,017                          & 0,239                          & \textbf{0,007} & 0,024                          & 0,244                          & \textbf{0,009} \\
                                                      & \textbf{RotatE}    & 0,012                          & 0,212                          & \textbf{0,005} & 0,017                          & 0,253                          & \textbf{0,005} & 0,026                          & 0,248                          & \textbf{0,007} \\
                                                      & \textbf{RESCAL}    & 0,011                          & 0,211                          & \textbf{0,001} & 0,018                          & 0,248                          & \textbf{0,002} & 0,027                          & 0,24                           & \textbf{0,004} \\
                                                      & \textbf{DistMult}  & 0,01                           & 0,208                          & \textbf{0,003} & 0,017                          & 0,248                          & \textbf{0,005} & 0,027                          & 0,242                          & \textbf{0,007} \\
                                                      & \textbf{ConvE}     & 0,006                          & 0,191                          & \textbf{0,001} & 0,011                          & 0,237                          & \textbf{0,001} & 0,018                          & 0,247                          & \textbf{0,002} \\
                                                      & \textbf{ComplEx}   & 0,009                          & 0,212                          & \textbf{0,002} & 0,017                          & 0,25                           & \textbf{0,003} & 0,025                          & 0,24                           & \textbf{0,005} \\ \midrule
\multirow{6}{*}{\textbf{FB15k}}                       & \textbf{TransE}    & 0,06                           & 0,277                          & \textbf{0,023} & 0,026                          & 0,166                          & \textbf{0,007} & 0,019                          & 0,106                          & \textbf{0,005} \\
                                                      & \textbf{RotatE}    & 0,039                          & 0,19                           & \textbf{0,014} & 0,023                          & 0,123                          & \textbf{0,011} & 0,019                          & 0,086                          & \textbf{0,009} \\
                                                      & \textbf{RESCAL}    & 0,028                          & 0,246                          & \textbf{0,002} & 0,024                          & 0,204                          & \textbf{0,003} & 0,025                          & 0,148                          & \textbf{0,005} \\
                                                      & \textbf{DistMult}  & 0,017                          & 0,116                          & \textbf{0,005} & 0,011                          & 0,095                          & \textbf{0,003} & 0,011                          & 0,08                           & \textbf{0,002} \\
                                                      & \textbf{ConvE}     & 0,02                           & 0,198                          & \textbf{0,005} & 0,019                          & 0,186                          & \textbf{0,007} & 0,023                          & 0,166                          & \textbf{0,009} \\
                                                      & \textbf{ComplEx}   & 0,013                          & 0,081                          & \textbf{0,004} & 0,01                           & 0,071                          & \textbf{0,003} & 0,01                           & 0,061                          & \textbf{0,003} \\ \midrule
\multirow{4}{*}{\textbf{CoDEx-S}}                     & \textbf{TransE}    & 0,049                          & 0,317                          & \textbf{0,007} & 0,084                          & 0,328                          & \textbf{0,007} & 0,11                           & 0,254                          & \textbf{0,007} \\
                                                      & \textbf{RESCAL}    & 0,049                          & 0,279                          & \textbf{0,001} & 0,081                          & 0,292                          & \textbf{0,002} & 0,117                          & 0,241                          & \textbf{0,004} \\
                                                      & \textbf{ConvE}     & 0,042                          & 0,273                          & \textbf{0,000} & 0,07                           & 0,273                          & \textbf{0,001} & 0,108                          & 0,232                          & \textbf{0,001} \\
                                                      & \textbf{ComplEx}   & 0,039                          & 0,245                          & \textbf{0,002} & 0,069                          & 0,265                          & \textbf{0,005} & 0,116                          & 0,253                          & \textbf{0,011} \\ \midrule
\multirow{2}{*}{\textbf{CoDEx-M}}                     & \textbf{ConvE}     & 0,011                          & 0,17                           & \textbf{0,001} & 0,022                          & 0,195                          & \textbf{0,001} & 0,044                          & 0,199                          & \textbf{0,002} \\
                                                      & \textbf{ComplEx}   & 0,036                          & 0,173                          & \textbf{0,004} & 0,06                           & 0,185                          & \textbf{0,007} & 0,09                           & 0,181                          & \textbf{0,012} \\ \midrule
\multirow{5}{*}{\textbf{CoDEx-L}}                     & \textbf{TransE}    & 0,078                          & 0,193                          & \textbf{0,071} & 0,04                           & 0,178                          & \textbf{0,027} & 0,043                          & 0,176                          & \textbf{0,019} \\
                                                      & \textbf{TuckER}    & 0,016                          & 0,149                          & \textbf{0,001} & 0,028                          & 0,158                          & \textbf{0,001} & 0,048                          & 0,161                          & \textbf{0,003} \\
                                                      & \textbf{RESCAL}    & 0,015                          & 0,143                          & \textbf{0,001} & 0,025                          & 0,154                          & \textbf{0,002} & 0,043                          & 0,159                          & \textbf{0,003} \\
                                                      & \textbf{ConvE}     & 0,011                          & 0,139                          & \textbf{0,001} & 0,022                          & 0,153                          & \textbf{0,002} & 0,04                           & 0,157                          & \textbf{0,004} \\
                                                      & \textbf{ComplEx}   & 0,013                          & 0,129                          & \textbf{0,003} & 0,023                          & 0,14                           & \textbf{0,004} & 0,04                           & 0,147                          & \textbf{0,006} \\ \midrule
\textbf{YAGO3-10}                    & \textbf{ComplEx}   & 0,104                          & 0,198                          & \textbf{0,029} & 0,111                          & 0,194                          & \textbf{0,035} & 0,075                          & 0,148                          & \textbf{0,027} \\ \bottomrule
\end{tabular}
\end{table*}

\end{document}